\documentclass{article}

\usepackage[accepted]{icml2022}

\usepackage{multicol}
\usepackage{microtype}
\usepackage{graphicx}
\usepackage{xurl,hyperref}
\usepackage[nolist,nohyperlinks]{acronym}
\usepackage{subcaption}
\usepackage[T1]{fontenc}
\usepackage{cleveref}
\usepackage{tabularray}

\begin{acronym}
  \acro{AI}{Artificial Intelligence}
  \acro{ML}{Machine Learning}
  \acro{FLOP}{floating point operations}
  \acro{CEG}{compute-equivalent gain}
\end{acronym}

\icmltitlerunning{AI capabilities can be significantly improved without expensive retraining}

\begin{document}

\twocolumn[
\icmltitle{AI capabilities can be significantly improved without expensive retraining}

\icmlsetsymbol{equal}{*}

\begin{icmlauthorlist}
    \icmlauthor{Tom Davidson}{openphil}
    \icmlauthor{Jean-Stanislas Denain}{ucberkeley}
    \icmlauthor{Pablo Villalobos}{epoch}
    \icmlauthor{Guillem Bas}{ORCG}
\end{icmlauthorlist}

\icmlaffiliation{epoch}{Epoch}
\icmlaffiliation{openphil}{Open Philantropy}
\icmlaffiliation{ucberkeley}{UC Berkeley}
\icmlaffiliation{ORCG}{Observatorio de Riesgos Catastróficos Globales}

\icmlcorrespondingauthor{Tom Davidson}{tom@openphilantropy.org}

\vskip 0.3in
]

\printAffiliationsAndNotice{}

\begin{abstract}
State-of-the-art AI systems can be significantly improved without expensive retraining via “post-training enhancements”—techniques applied after initial training like fine-tuning the system to use a web browser. We review recent post-training enhancements, categorizing them into five types: tool-use, prompting methods, scaffolding, solution selection, and data generation. Different enhancements improve performance on different tasks, making it hard to compare their significance. So we translate improvements from different enhancements into a common currency, the compute-equivalent gain: how much additional training compute would be needed to improve performance by the same amount as the enhancement. Our non-experimental work shows that post-training enhancements have significant benefits: most surveyed enhancements improve benchmark performance by more than a 5x increase in training compute, some by more than 20x. Post-training enhancements are relatively cheap to develop: fine-tuning costs are typically <1\% of the original training cost. Governing the development of capable post-training enhancements may be challenging because frontier models could be enhanced by a wide range of actors.

\end{abstract}

\section{Executive summary}
\label{sec1}

It is important to understand the drivers of AI progress, to inform both forecasts of future progress and initiatives for AI governance. Previous analyses have mostly focused on the initial development of an AI system, called \textit{pre-training}: computationally expensive training runs where models learn from massive amounts of data. Researchers have measured how quickly the \textit{inputs} to training runs have been increasing -- more computational resources  (``compute'')
\cite{sevilla2022}, more data \cite{Villalobos2022}, better algorithms \cite{Hernandez2020, erdil2023algorithmic} -- and how bigger training runs translate into improved performance on downstream tasks \cite{Srivastava2023}.

There has been little focus in the AI forecasting literature on what we call ``post-training enhancements'': techniques for improving the performance of an AI model after it is initially developed.\footnote{Many papers introduce and evaluate individual enhancements, but there is little work systematically evaluating multiple enhancements.
\cite{Villalobos2023} discuss techniques for improving performance at the cost of more expensive inference, which count as post-training enhancements under our definition. We focus on a wider set of enhancements, most of which do not make inference much more expensive. \cite{Anderljung2023} briefly discusses various ``post-deployment enhancements'', which are a subset of post-training enhancements.}
Post-training enhancements are commonly used in deployed AI systems. Examples include: teaching the model to browse the web, asking the model to ``think step-by-step'', fine-tuning\footnote{Fine-tuning refers to additional training that occurs after pre-training, often to elicit specific skills and tendencies from the model. Fine-tuning typically uses a much smaller quantity of data than pre-training.}
on task-specific curated data sets, and using the model to power an autonomous AI agent (e.g. AutoGPT).\footnote{See \citet{Mialon2023} for a survey of some of these enhancements, which they call augmentations.}

\begin{figure}[htb]\centering
\includegraphics[width=0.45\textwidth]{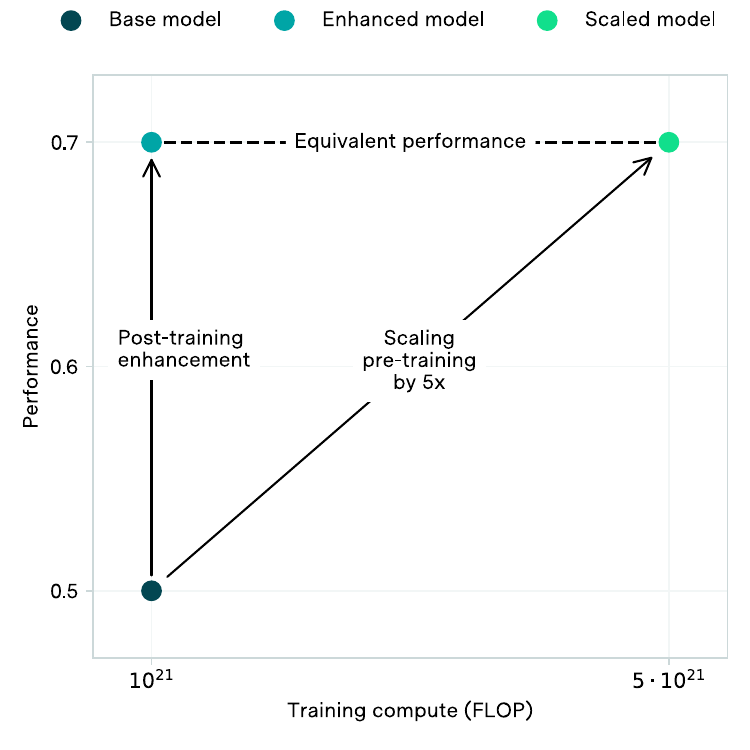}
\caption{Illustration of the compute-equivalent gain. The enhanced model has the same performance as a non-enhanced model trained with $5\times$ more compute.}
\label{fig:toy}
\end{figure}

It is hard to meaningfully compare the benefits of post-training enhancements that apply to different domains. For example, how does 10\% greater accuracy on the MATH benchmark \cite{Hendrycks2021} compare to 10\% greater accuracy in a multiple choice knowledge test, or to 10\% lower perplexity in a language modeling task? To address this problem, we translate performance gains on different benchmarks into a common currency. In particular, we estimate \textbf{how much additional training compute would have been needed to improve benchmark performance by as much as the post-training enhancement}.\footnote{Assuming that the additional training compute is used to train further on the pre-training distribution, rather than on some task-specific dataset.}
We call this the ``compute-equivalent gain'' (CEG). In the toy example below, the post-training enhancement improves performance by the same amount as increasing the training compute by $5\times$; so the CEG is 5 (see \Cref{fig:toy}).

\begin{figure*}[htb]\centering
\includegraphics[width=0.9\textwidth]{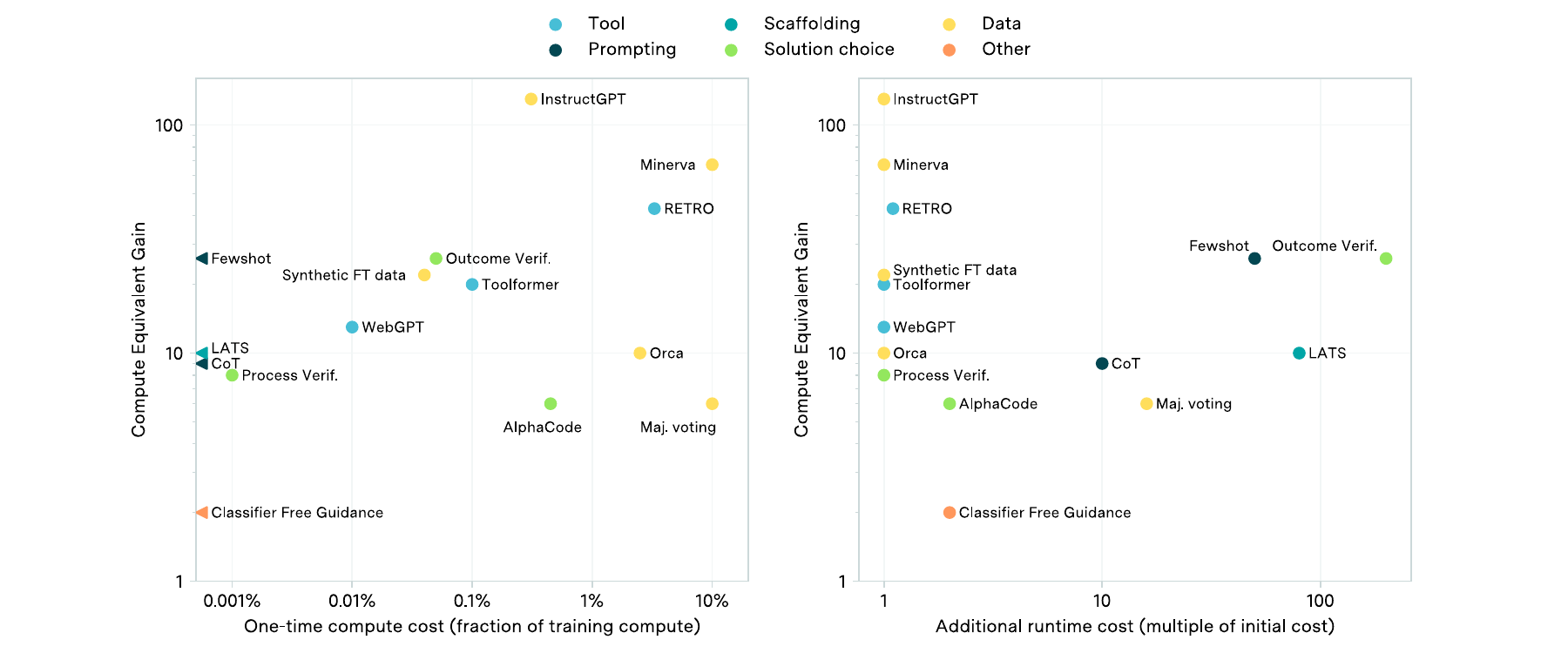}
\caption{Distribution of CEG and additional costs of the techniques we studied. The one-time cost is given as a fraction of pre-training, the runtime cost is relative to the runtime cost without the enhancement. Enhancements without one-time cost are shown with an arrow on the y axis.}
\label{fig:main}
\end{figure*}

A CEG of 5 for a particular benchmark doesn't necessarily imply that the post-training enhancement is as useful \textit{overall} as increasing the training compute by $5\times$. Post-training enhancements are often, though not always, domain-specific (e.g. learning to use a calculator), whereas the performance gains from additional training compute tend to be more general.\footnote{The fact that pre-training helps in a wide variety of tasks has been widely documented \cite{Brown2020, Srivastava2023}. In particular, pre-training scaling usually also improves performance in the tasks in which each particular post-training enhancement claims benefits (see, for example, \cite{Press2022} and
\cite{Wei2022} for chain-of-thought prompting, \cite{Lewkowycz2022} for data cleaning, \cite{Parisi2022} for tool use, and \cite{Li2022} for solution choice).}
As we illustrate in this paper, the CEG is a useful way to get a \textit{rough} sense of the significance of an enhancement.

We provide an overview of recent post-training enhancement techniques. We are not exhaustive, but rather present a broadly representative set of examples. We divide post-training enhancements into five categories:\footnote{Note that some of these categories often overlap. For example, scaffolding enhancements commonly also generate multiple candidate solutions.}
\begin{enumerate}
\item \textbf{Tool enhancements:} teaching an AI system to use new tools, like a web browser.

\item \textbf{Prompting enhancements:} changing the text-based input to the model to steer its behavior and reasoning, e.g. including an example response to a similar question.

\item \textbf{Scaffolding enhancements:} programs that structure the model's reasoning and the flow of information between different copies of the model (e.g. producing AI agents).

\item \textbf{Solution choice enhancements:} techniques for generating and then choosing between multiple candidate solutions to a problem.

\item \textbf{Data enhancements:} techniques for generating more, higher-quality data for fine-tuning.
\end{enumerate}

We quantify the benefits of post-training enhancements by calculating the CEG. We also estimate each enhancement's costs: the one-time cost of compute for fine-tuning the model to be able to make use of the enhancement, and any increased inference cost associated with the enhancement (\Cref{fig:main}).

Most of the post-training enhancements we studied have a CEG above 5, and many are above 20. This corresponds to significant performance improvements in particular domains; improvements that would have been very expensive to achieve via spending more on pre-training. We find that the fine-tuning costs are low: typically $<1\%$ and sometimes $<0.01\%$ of the pre-training cost. The inference costs vary: often there is no additional inference cost and other times inference becomes $100\times$ more expensive. Post-training enhancements with a higher CEG tend to have higher costs in either fine-tuning or inference, which is unsurprising given the performance benefits of additional training and inference compute \cite{Villalobos2023}. There are development costs we do not estimate, like human labor.

We did not conduct our own experiments to measure the CEG, but used results from other research papers. This made it challenging to get exact estimates, and there were sometimes confounding factors (see \cref{sec4}). For this reason, we don't have high confidence in each individual CEG estimate, but we think that in aggregate they are informative about the typical benefit produced by an enhancement.

We predict that cheap post-training enhancements will continue improving model capabilities. Researchers have combined together multiple different post-training enhancements for increased benefit, for example by using both majority vote and a new technique for data cleaning \cite{Lewkowycz2022}. Many post-training enhancements improved upon previous versions; for example, the LATS agent outperforms previous agents and ``chain of thought'' prompting has since been improved upon in multiple papers \cite{Wang2023, Press2022, Huang2022}.
Often post-training enhancements only become effective when models are sufficiently large (e.g. \cite{Wei2022, Schick2023}). These patterns suggest that researchers will continue to improve frontier AI capabilities by finding more post-training enhancements and combining them together in new ways. However, it's unclear how the CEG of these enhancements scales with training compute, and how much total improvement they can provide when combined. Further experimental research could shed light on this important open question.

Post-training enhancements are already enabling new and beneficial applications of AI, allowing systems to be specialized to particular use cases.\footnote{For example, \href{https://www.tryklarity.com/}{Klarity} uses post-training enhancements to help automate accounting work.} However, post-training enhancements could enable some dangerous applications. Firstly, they could make a model more \textit{generally capable} at both benign and illicit tasks, e.g. creating a generalist agent that could be used to write phishing emails \cite{kinniment2023}. Secondly, they could enhance capabilities in a \textit{narrow but potentially dangerous} \textit{domain}, e.g. connecting a model to robotic hardware for synthesizing chemicals \cite{Boiko2023}.

Post-training enhancements pose a distinctive governance challenge. Training a frontier AI model today is expensive -- the cost of compute for training GPT-4 is estimated to be around \$50 million \cite{epoch2023aitrends}. This means that few actors can advance frontier capabilities via pre-training. By contrast, post-training enhancements are often cheap. The compute cost of fine-tuning is sometimes $<0.01\%$ the cost of pre-training and, though post-training enhancements can increase inference costs, the inference costs for a single user are still extremely low compared to the cost of pre-training. So a wider range of actors could expand the capabilities of frontier AI models via post-training enhancements, potentially in unanticipated ways. 

When evaluating a model's safety for release, it may be necessary to consider not only its current capabilities but also those that could be unlocked by future post-training enhancements. Recognizing the potential for these enhancements to significantly raise a model's dangerous capabilities, it could be prudent to use a  'safety buffer', as discussed in \Cref{sec6}, where protective measures are activated at capability levels lower than those presently identified as hazardous.

The paper is structured as follows:
\begin{itemize}
\item \Cref{sec2} lays out our conceptual framework for quantifying the benefits (via the CEG) and compute costs of post-training enhancements.

\item \Cref{sec3} applies this framework to analyze fourteen examples of post-training enhancements from the literature, and discusses many others.

\item \Cref{sec4} describes limitations of our CEG estimates.

\item \Cref{sec5} argues that post-training enhancements will continue to improve capabilities in the future, that they can often be combined, and that they can be either narrow or general.

\item \Cref{sec6} considers policy implications, especially relating to dangerous capabilities.

\item \Cref{sec7} presents possible directions for future work.

\item \Cref{sec8} concludes.
\end{itemize}

Our core contribution is analyzing the benefits (via the CEG) and the compute costs of a wide range of post-training enhancements. \Cref{tab:main} summarizes our analysis.

\onecolumn

\begin{longtblr}[
caption = {Summary of post-training enhancements (PTEs), their compute-equivalent gain (CEG) and their associated costs.},
label = {tab:main}
]{colspec={X[0.7,l,m]X[1.3,l,m]X[1.1,l,m]X[1,c,m]X[1,c,m]},
rows={font=\small},
row{1}={font=\bfseries\small},
width=\textwidth,
colsep=2.5pt,
rowsep=3pt,
rowhead = 1,
}
\hline \SetCell{c,m} Technique &\SetCell{c,m} Explanation &
\SetCell{c,m} Compute-equivalent gain (CEG) & \SetCell{c,m} One-time cost (fraction of training compute) & \SetCell{c,m} New inference cost (multiple of initial cost)
\\ \hline
\SetCell[c=5]{l,m,font=\itshape}Tool enhancements & & & &
\\ \hline
\hyperref[toolformer]{Toolformer} & Fine-tune a model to use a calculator, a Q\&A system, a search engine, a translation system, and a calendar. &$>20$ in benchmarks for factual knowledge,
math, and temporal questions
&$\sim 0.1 \%$ & $1$
\\
\hyperref[webgpt]{WebGPT}&
Fine-tune a model to use a web browser to answer factual questions and
provide citations
&$>13$ on a question-answering benchmark.

Larger gains when
combined with best-of-$n$
&$\sim 0.01 \%$ & $1$
\\
\hyperref[retro]{Memory retrieval}& The model retrieves text that is similar to the text it is predicting, and uses it to inform its predictions.
&$>43$ on one next-word-prediction dataset; lower for others.
& $<3.3 \%$ & $<1.1$
\\ \hline
\SetCell[c=5]{l,m,font=\itshape}
Prompting enhancements &&&&
\\ \hline
\hyperref[fewshot]{Few-shot prompting}
&Provide a few solved examples to the model
&$>26$ in SuperGLUE; lower on other benchmarks & $0$ & $>5$ and $<50$ in three examples
\\
\hyperref[cot]{Chain of thought}
&Encourage a model to think problems through ``step by step''.
&$>9$ on many benchmarks & $0$ & $10$
\\ \hline
\SetCell[c=5]{l,m,font=\itshape}
Scaffolding enhancements &&&&
\\ \hline
\hyperref[tot]{Tree of thoughts}
&The model generates thought candidates at each step, evaluates their
progress heuristically, and uses search algorithms to navigate the thought tree.
&Not enough data to measure the CEG.\smallskip

Better at deduction problems with multiple solution paths.
& 0 & $100$
\\
\hyperref[parsel]{Parsel} &
The model decomposes a complex task into natural language function
descriptions, generates modular implementations
for each, and searches over combinations of these implementations by testing against constraints.
&Not enough data to measure the CEG.\smallskip

Better at programming tasks: $75 \%$ higher pass rate than directly sampling AlphaCode or Codex.
\\
\hyperref[agents]{Agents} & A model assigns sub-tasks to copies of itself, reads and writes to memory, has a chance to learn from their mistakes, etc.
&$>10$ at HumanEval, comparing the LATS agent with basic few-shot prompting.
& 0 &$\sim 80\times$ (for LATS at HumanEval)
\\ \hline
\SetCell[c=5]{l,m,font=\itshape}
Solution choice enhancements&&&&
\\ \hline
\hyperref[verification]{Verification} & A verifier rates 100 candidate solutions and submits the one with the highest rating. &$>26$ on a math benchmark.
& $\sim 0.05 \%$ & $200$
\\
\hyperref[verification_proc]{Verification with process-based feedback}
&
Improves on a ``outcomes-based'' verifier baseline by fine-tuning a verifier with ``process-based'' feedback.
&$\sim8$ on a math benchmark
&  $\sim 0.001 \%$ & 1
\\
\hyperref[alphacode]{AlphaCode sample selection} &
Six techniques for choosing which coding solutions to submit out of 1000 s of candidates.
&$\sim 6$ on a coding benchmark
& $\sim 0.45 \%$ & $<2$
\\ \hline
\SetCell[c=5]{l,m,font=\itshape}
Data enhancements&&&&
\\ \hline
\hyperref[minerva]{Data cleaning} & Fine-tune a large language model on a carefully cleaned STEM dataset.
&$\sim 5$ in STEM benchmarks;\smallskip

$>67$ in math benchmarks
&$\sim 10 \%$ & $1$
\\
\hyperref[minerva]{Data cleaning + majority voting} & As above, but also generate multiple solutions and submit the most common answer.
&$>6$ in two STEM benchmarks.\smallskip

$>2000$ in two math benchmarks
&$\sim 10 \%$ & $>16$ for STEM benchmarks

$>64$ for math benchmarks
\\
\hyperref[orca]{Learning from a teacher model}
&Fine-tune a small model on detailed explanations produced by a larger model
&$\sim 10$ on a range of benchmarks
&$\sim 2.5 \%$ & 1
\\
\hyperref[syndata]{Generating your own fine-tuning data}&
Models write coding puzzles and solutions; solutions are automatically
checked; fine-tune on correct solutions
&$>22$ in a coding benchmark, compared to a baseline with no
fine-tuning for coding
&$\sim 0.04 \%$ & 1
\\
\hyperref[instructgpt]{Instruct GPT} &
Fine-tune a model on examples of humans following instructions;\smallskip

fine-tune against a reward model trained to predict
human preferences.
&$>3900$ at instruction following;\smallskip

$>130$ on some other NLP benchmarks;\smallskip

no gain on many NLP benchmarks
& $\sim 0.31 \%$ & $1$
\\ \hline

\end{longtblr}

\begin{multicols}{2}

\section{Conceptual framework}
\label{sec2}
\subsection*{Benefits of post-training enhancements}

Measuring the performance improvement from a post-training enhancement is not straightforward: there are hundreds of different benchmarks for AI capabilities, and different benchmarks often use different units to measure performance.\footnote{For example, language modeling benchmarks commonly use perplexity or cross-entropy, whereas question-answering or problem-solving benchmarks tend to use accuracy (see \cite{Srivastava2023} or \cite{Hoffmann2022} for some examples).}
A 5\% improvement in one benchmark may reflect a greater capability gain than a 5\% improvement in another. This makes it hard to compare the performance gains from different enhancements, when they improve performance on different benchmarks.

To address this problem, we would like to translate all performance gains into a common unit. A good candidate unit is training compute. The literature on scaling laws has identified a strong relationship between the compute spent on training and the general capabilities of ML models, so training compute is a good proxy for capability \cite{Henighan2020, Hoffmann2022, Srivastava2023}.

Concretely, we can calculate the additional compute that would be required to match the performance improvement from an enhancement. We call this quantity the compute-equivalent gain (CEG).\footnote{Note that the exact relation between training compute and capabilities depends on the model family, so our definition of compute-relative gain is also relative to a model family. See \Cref{apA} for more details.} It can be calculated in the following manner:
\begin{enumerate}
\item Measure the performance $p$ of an AI model trained with training compute $C$ \textit{without} the post-training enhancement. The model should be trained compute-optimally, so that $C$ is the minimal compute required to attain performance $p$.

\item Measure the performance $p*$ of an AI model trained optimally with training compute $C$ \textit{with} the post-training enhancement.

\end{enumerate}

\end{multicols}
\twocolumn

\begin{enumerate}
\setcounter{enumi}{2}
\item Estimate the training compute $C'$ needed to achieve performance $p*$ \textit{without} the post-training enhancement, again training optimally.

\item The CEG is given by $C'/C$
\end{enumerate}

A similar approach was used by \cite{Hilton2023} in the context of scaling laws. They define the ``intrinsic performance'' of a model as the minimum amount of training compute that would be required to reach the same performance obtained by the model.\footnote{The CEG can be seen as the ratio of the intrinsic performances of two models, one in the family of enhanced models and the other in the original family of models.}

Since we don't have the capacity to run our own evaluations and have to rely on reported evaluation data, we often can't calculate the CEG exactly. Instead, we often calculate a lower bound, which can be done if the evaluation data contains a model with the post-training enhancement that outperforms some bigger model without the enhancement.\footnote{See \Cref{apA} for more details on the problems associated with these bounds and possible solutions.}

Gains from enhancements and gains from scaling might interact and compound in nontrivial ways. For example, chain-of-thought prompting typically improves performance more in bigger models. We don't attempt to quantify these effects. We calculate the CEG at the scale used in the paper that introduces the post-training enhancement. We calculate the CEG using a reproducible methodology -- see \cref{apA} for a detailed description.

\subsection*{Cost of post-training enhancements}

Once again, quantifying the cost of enhancing a model is not trivial. There are many factors that influence cost: the technical labor needed to incorporate the technique, the cost of the hardware, the effect on inference latency, etc \cite{cottier2023}. We will focus on the compute costs of these enhancements because these are relevant for certain governance questions (see \Cref{sec6}) and can be readily calculated.

Firstly, we calculate a one-time compute cost for fine-tuning. Even if the post-training enhancement is not centrally about adapting a model to a downstream task, fine-tuning may be needed to train a verifier, teach an AI to use a new tool, or to add on a memory retrieval mechanism. Secondly, we calculate any ongoing higher inference costs from an enhancement.

The economic relevance of these costs depends on the use case. For example if one is primarily concerned with demonstrating a certain capability, the one-time training costs will be more relevant because inference costs for a single use are typically very low. Meanwhile, inference costs will be more relevant for large-scale deployment due to the large number of inferences that will be performed.

\section{Analysis of post-training enhancements}
\label{sec3}

In this section we examine the performance gains from many post-training enhancements, each time estimating benefits (via the CEG), the one-time cost of compute for fine-tuning, and any increased inference cost associated with the enhancement. \Cref{tab:main}, at the end of \Cref{sec1}, summarizes the analysis.

\subsection*{Tool enhancements}

The post-training enhancements in this section give models access to tools to improve their performance on downstream tasks. ChatGPT can already use many such tools via various plugins.

\subsubsection*{Toolformer}
\label{toolformer}

\cite{Schick2023} augment a model with tools to improve performance on a variety of downstream tasks. They introduce Toolformer, a 6.7B parameter language model fine-tuned to use a number of tools: a calculator, a Q\&A system, a search engine, a translation system, and a calendar. We estimate that the compute used for fine-tuning is $\sim0.1\%$ of the compute used for pre-training.\footnote{Toolformer is based on GPT-J, which took 1.5e22 FLOP to train \cite{Komatsuzaki2021}. Toolformer was trained for 2000 steps, with a batch size of 128, and sequence length of 1024 (Appendix B of \citet{Schick2023}). Since the model size is 6B, we estimate the fine-tuning process took $\mathrm{2000\ast128\ast1024\ast6e9\ast6 \approx1e19}$ FLOP.}

Toolformer uses different tools for different downstream tasks and the performance gain varies by task.

\begin{itemize}
\item Toolformer outperforms the 175B GPT-3 model on benchmarks for \textit{factual knowledge} (table 3), \textit{math} (table 4) and \textit{temporal questions} (table 7). We estimate GPT-3 is trained with $20\times$ more compute than Toolformer,\footnote{As stated in a previous footnote, we estimate Toolformer was trained with 1.5e22 FLOP. This contrasts with an estimated 3.1e23 FLOP for GPT-3 (see \citet{Schick2023}).}
so the CEG is $>20$.

\item Toolformer outperforms the 66B OPT model on benchmarks for \textit{question answering} (table 5). We estimate OPT is trained with $7\times$ more compute than Toolformer,\footnote{As stated in a previous footnote, we estimate Toolformer was trained with 1.5e22 FLOP. Meanwhile, OPT-66B was trained for 140k steps, using a batch size of 2M tokens (see the \href{https://github.com/facebookresearch/metaseq/blob/main/projects/OPT/chronicles/OPT_Baselines_Logbook.pdf}{OPT baselines logbook} and Table 1 in \citet{Zhang2022}, respectively), so training took $\mathrm{140e3\ast 2e6\ast 66e9\ast 6 = 1.1e23}$ FLOP. The compute gain is then $\mathrm{1.1e23 / 1.5e22 \approx 7}$.}
so the CEG is $>7$.

\item Toolformer doesn't show significant improvement in \textit{translation} (table 6) despite having access to a translator.
\end{itemize}

So the CEG ranges from 1 (no gain) to $>20$ depending on the downstream task.

\subsubsection*{WebGPT}
\label{webgpt}

\citet{Nakano2021} train language models to use a web browser to answer long-form questions more accurately. They equip GPT-3 with access to a web browser, and collect demonstrations of humans using the web browser tool to answer questions from the ELI5 dataset. They then fine-tune the GPT-3 model on those demonstrations. We estimate that the compute used for fine-tuning is $\sim0.01\%$ of the compute used for pre-training.\footnote{The dataset consisted of 6209 demonstrations, and training ranged from 2 to 5 epochs (see Appendixes B and E of \citet{Nakano2021}). Assuming an average of 800 tokens per demonstration (obtained by scraping 100 demonstrations from the database), we get 10 to 25 million tokens seen during training. In comparison to the 300B tokens seen during pretraining (see \citet{Brown2020}), this requires $\sim 0.01\%$ as much compute. }

\begin{figure}[htb]\centering
 \includegraphics[width=0.45\textwidth]{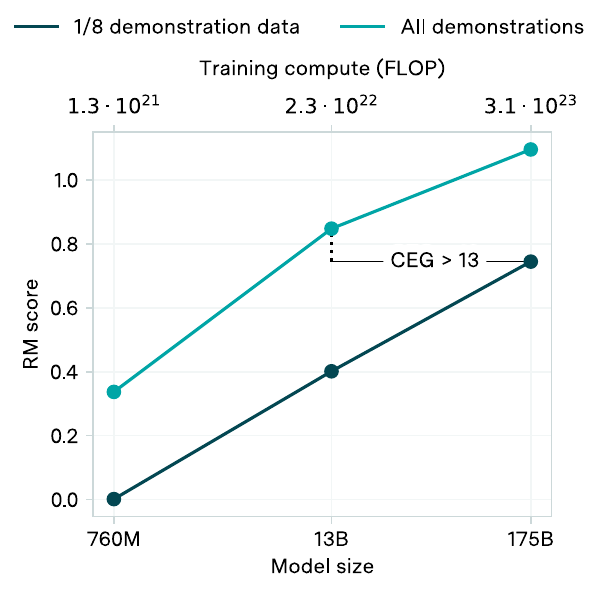}
 \caption{Performance of WebGPT with different quantities of demonstrations used for fine-tuning.\protect\footnotemark}
 \label{fig:webgpt}
\end{figure}

\Cref{fig:webgpt} shows the scaling trends with fine-tuning data. It can be seen that increasing the fine-tuning dataset by $8\times$ leads to more improvement than scaling model size from 13B to 175B.\footnote{Increasing the size of the fine-tuning dataset by a larger amount would probably lead to larger improvements.}
We estimate that this corresponds to a CEG of $\sim 13\times$.\footnote{All GPT-3 models were trained on 300B tokens (see Table D.1 in \citet{Brown2020}), so the parameter increase is proportional to the compute increase.}

\footnotetext{Data from Figure 6 of \citet{Nakano2021}.}
In addition, \citet{Nakano2021} collect human comparisons between different model-generated responses to questions, and train a reward model on those comparisons. This reward model is used to perform reinforcement learning and best-of-n sampling -- a solution choice enhancement where multiple candidate solutions are generated and the one with the highest score (according to the reward model) is submitted. We estimate that the compute used for training the reward model is $\sim0.01\%$ of the compute used for pre-training.\footnote{The dataset consisted of 21548 demonstrations, and training took 1 epoch (see Appendixes B and E of \citet{Nakano2021}). Assuming an average of 1600 tokens per comparison (two demonstrations), we get 35 million tokens seen, which again corresponds to about 0.01\% of the pre-training compute.}

\begin{figure}[htb]\centering
 \includegraphics[width=0.45\textwidth]{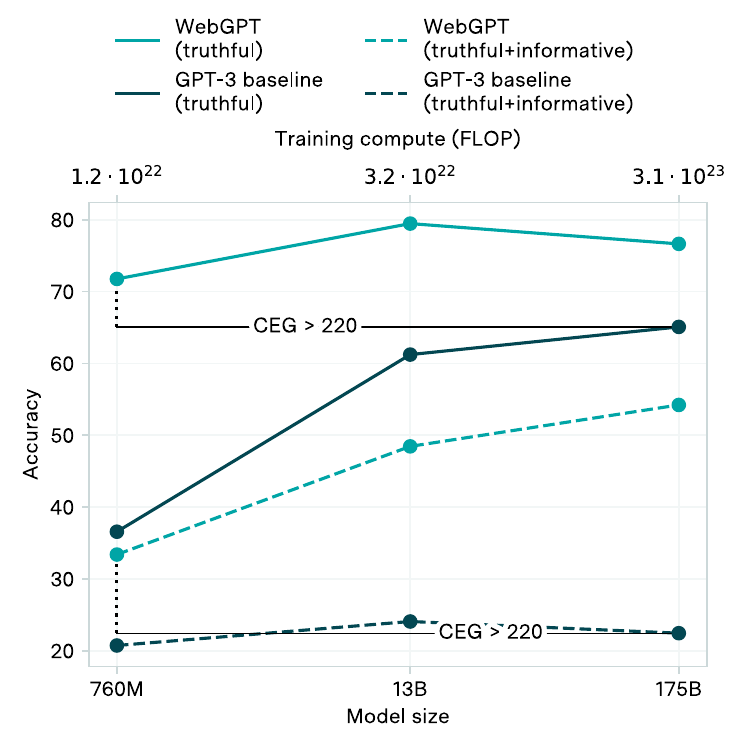}
 \caption{Performance of WebGPT and GPT-3 on TruthfulQA.\protect\footnotemark}
 \label{fig:webgpt_truth}
\end{figure}

They compare GPT-3 with WebGPT + best-of-$n$ on the benchmark TruthfulQA. The performance of the 760M model -- on both ``\% truthful'' and ``\% truthful and informative'' -- improves more from adding WebGPT than from moving to the 175B model (which has $220\times$ more training compute, see \Cref{fig:webgpt_truth}). So the CEG is $>220$.

\footnotetext{Data from Figure 3 of \citet{Nakano2021}.}

\subsubsection*{Memory retrieval}
\label{retro}

\citet{Borgeaud2021} add a text retrieval mechanism to a pre-trained LLM via fine-tuning. They call this process "RETROfitting". Retrieval mechanisms work by storing a large number of short text sequences in a database. When the model is predicting a new token, the current text is compared to the stored text and the most similar stored text sequences are selected.\footnote{The nearest neighbors are computed using the embeddings produced by a frozen BERT model.}
The information from these stored sequences is used to inform predictions of upcoming tokens. This architecture is illustrated in the following figure.

\begin{figure}[htb]\centering
\includegraphics[width=0.45\textwidth]{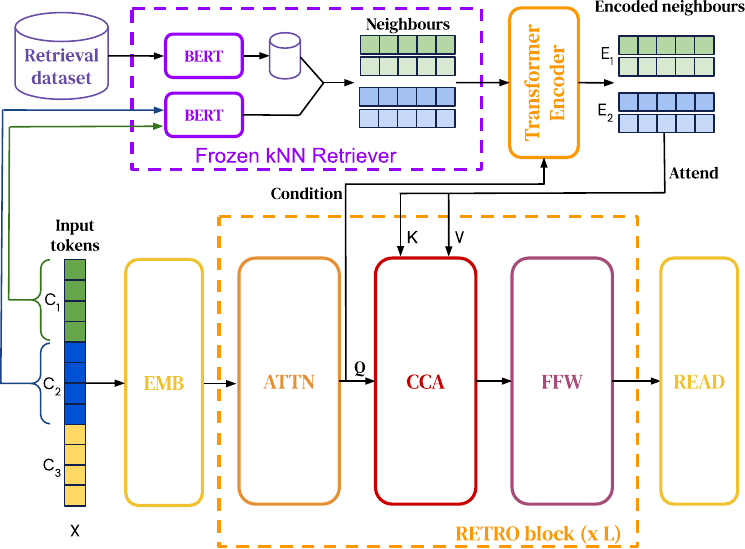}
\caption{RETRO architecture. Extracted from
\citet{Borgeaud2021}.}
\label{fig:retro_arch}
\end{figure}

\begin{figure*}[htb]\centering
 \includegraphics[width=\textwidth]{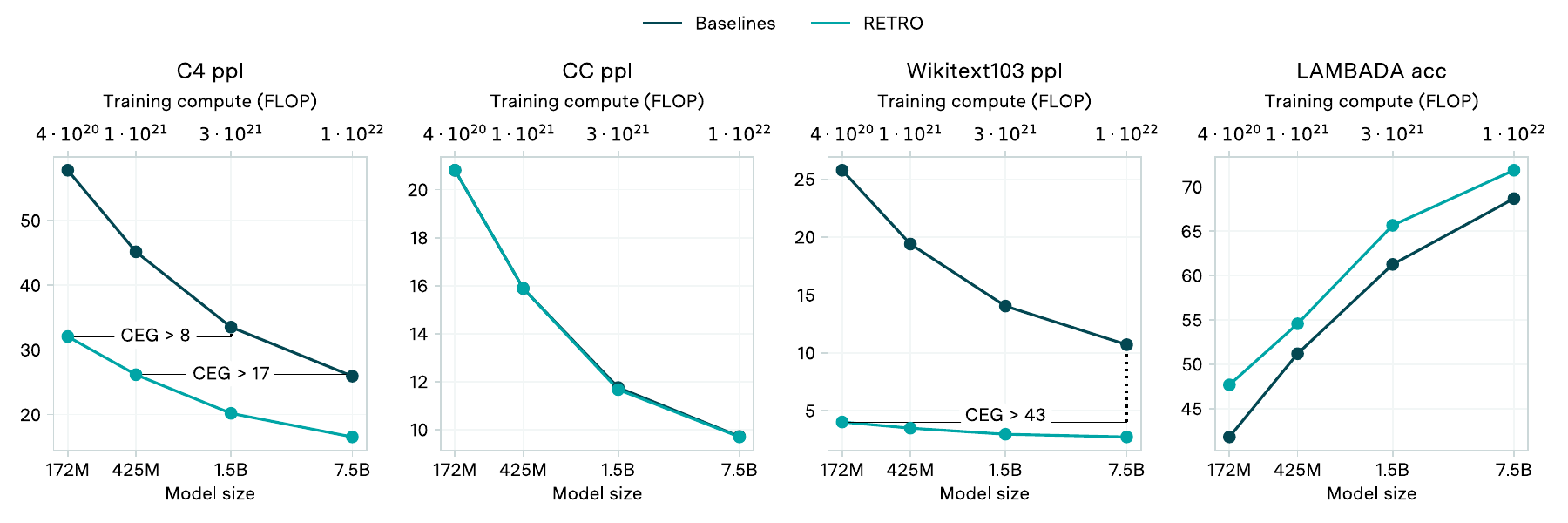}
 \caption{Performance of ``RETRO-fitted'' models on four benchmarks. Note that C4, CC and Wikitext103 measure performance in perplexity (ppl, lower is better), while LAMBADA measures it in accuracy (acc, higher is better).\protect\footnotemark}
 \label{fig:retro}
\end{figure*}
\footnotetext{Data from Figure 5 of \citet{Borgeaud2021}.}

While retrieval models can be trained from scratch, here we are interested in adding a retrieval mechanism to a pre-trained model. This can be done with some additional fine-tuning. In the case of RETRO, the fine-tuning compute is $<3.3\%$ of the pre-training compute.\footnote{The additional weights of the retrieval mechanism are less than 10\% of the weights of the original model, and the model is fine-tuned on 3\% of the original pre-training data. Therefore, the compute used in fine-tuning is less than $0.03\ast 1.10 = 3.3\%$.}
The retrieval mechanism increases the cost of inference by less than 10\%.\footnote{The additional weights of the retrieval mechanism are less than 10\% of the weights of the original model. The cost of computing nearest neighbors is amortized over many tokens and therefore we ignore it. }

The results can be seen in \Cref{fig:retro}. The improvement varies by dataset:
\begin{itemize}
\item In C4 (far left), the 172M-parameter fine-tuned model  achieves lower perplexity than the 1.5B-parameter baseline. In addition, the 425M model outperforms the 7.5B baseline. This corresponds to CEGs of 8 and 17.\footnote{All models were trained on 420B tokens (see Appendix C.1 of \citet{Borgeaud2021}), so the training compute is proportional to the number of parameters. 1.5B / 172M $\approx 8$ and 7.5B / 425M $\approx 17$. }

\item In Curation Corpus and LAMBADA there is almost no improvement: the accuracy of the fine-tuned models is close to their respective baselines.

\item In Wikitext103 all the RETRO fine-tuned models significantly outperform all the baselines. This corresponds to a CEG of 43.\footnote{All models were trained on 420B tokens (see Appendix C.1 of \citet{Borgeaud2021}), so the training compute is proportional to the number of parameters. $\mathrm{7.5B / 172M \approx 43}$. }
\end{itemize}

\subsection*{Prompting enhancements}

\subsubsection*{Few-shot prompting}
\label{fewshot}

In one of the earliest examples of prompting innovations, \citet{Brown2020} prompt large language models with a series of solved examples of the task. While smaller models don’t benefit much from adding these examples, larger models are able to learn how to perform the task solely from context, without any fine-tuning.

\begin{figure*}[htb]\centering
 \includegraphics[width=\textwidth]{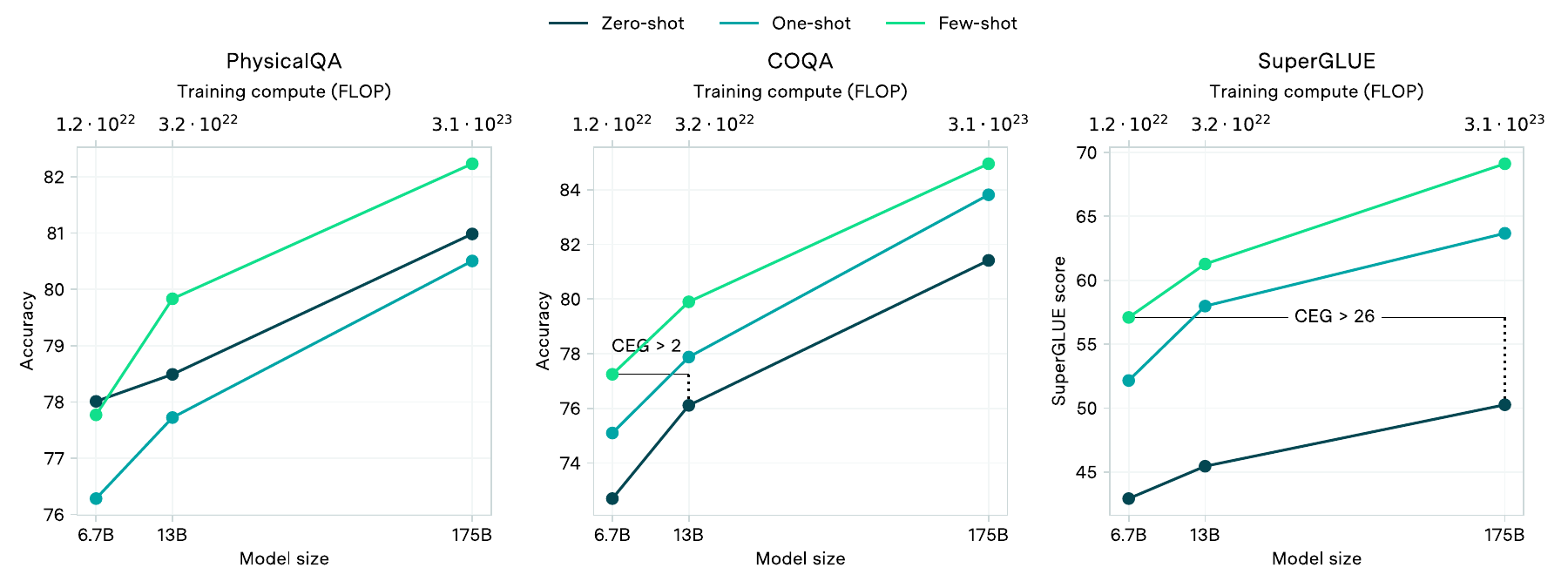}
 \caption{Performance of GPT-3 with few shot prompting on three benchmarks.\protect\footnotemark}
 \label{fig:fewshot}
\end{figure*}

The gains vary by task (see \Cref{fig:fewshot}): in PhysicalQA and COQA, a 6.7B-parameter model using few-shot prompting outperforms a 13B-parameter model with one-shot and zero-shot prompting, respectively. This corresponds to a CEG of $\sim 2$. Meanwhile, in SuperGLUE the gain is more significant: the 6.7B-parameter model with one-shot and few-shot prompting outperforms the 175B-parameter model with zero-shot prompting, corresponding to a CEG $>26$.\footnote{All GPT-3 models were trained on 300B tokens, so their training compute is proportional to the model size (Table D.1 of \citet{Brown2020}). 13B/6.7B $\approx 2$,  175B / 6.7B $\approx 26$.}

\footnotetext{Data from figures 3.6, 3.7 and 3.8 of  \citet{Brown2020}.}

\subsubsection*{Chain of thought}
\label{cot}

\citet{Wei2022} improve the reasoning ability of large language models by prompting them to generate a ``chain of thought'' (CoT), i.e. to think through their reasoning step-by-step. Language models have limited performance in tasks which require several serial steps of reasoning, like arithmetic or logical tasks. The authors mitigate this problem by including a few examples of correct reasoning processes in the language model's context window. The model then uses a similar reasoning process to answer subsequent questions. The method requires no additional training and doesn't specialize the model in any particular task.

While this post-training enhancement does not require any additional training compute, it usually increases the compute usage per inference. We estimate that the inference cost increases by up to $10\times$.\footnote{Based on the ratio between the length in tokens of the chains of thought and the answers themselves, in the prompts provided by the authors. The chains of thought are up to $10\times$ longer than the answers (see Appendix G of \citet{Wei2022}).}

They test the approach in three domains: arithmetic reasoning, commonsense reasoning, and symbolic reasoning. The improvement from chain-of-thought usually increases with the scale of the model.

\begin{figure*}[htb]\centering
 \includegraphics[width=\textwidth]{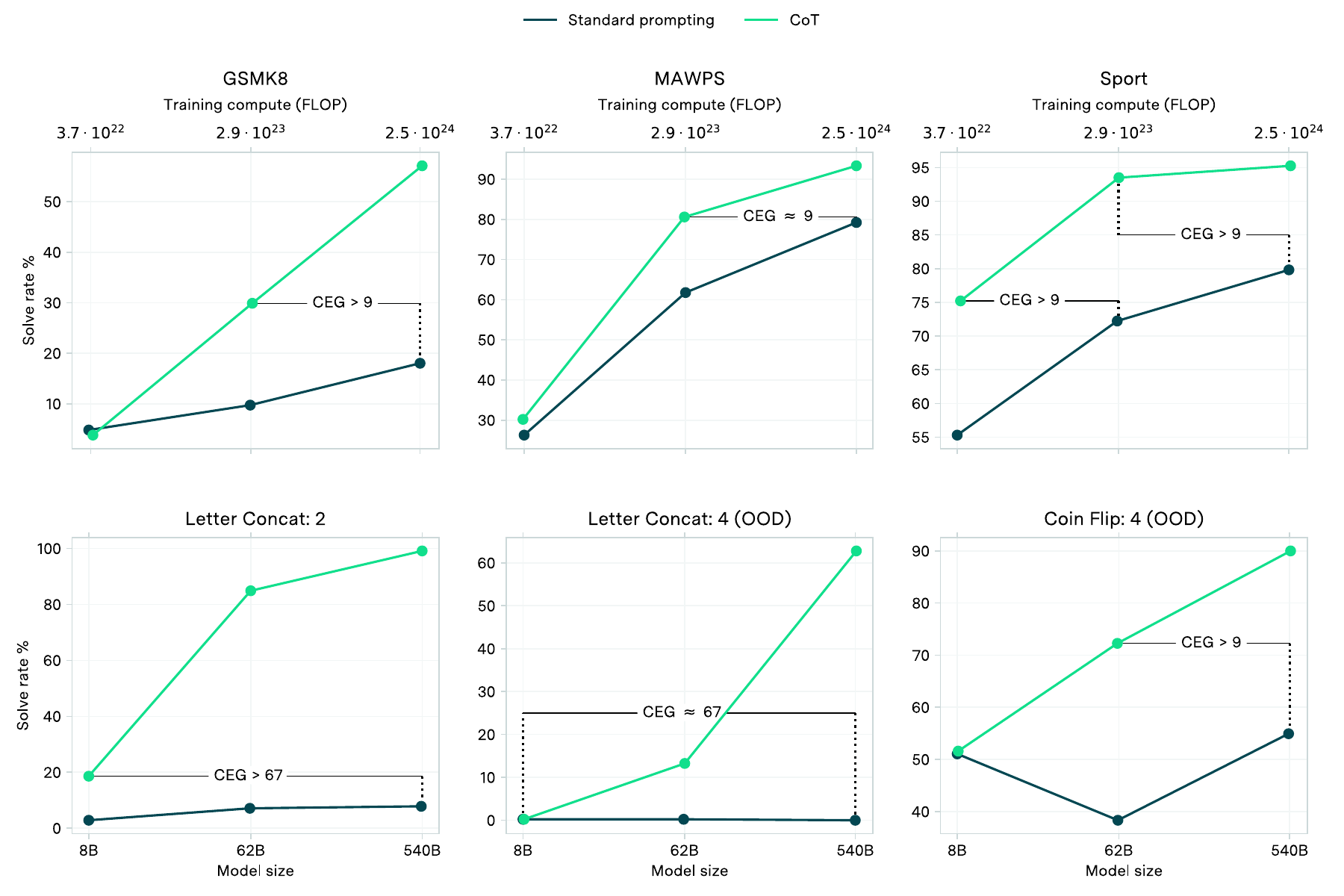}
 \caption{Performance of Chain of Thought and standard prompting for PaLM in several benchmarks.\protect\footnotemark}
 \label{fig:cot}
\end{figure*}
\footnotetext{Data from figures 4, 7 and 8 of \citet{Wei2022}.}

The performance gain varies significantly by task, as can be seen in \Cref{fig:cot}:

\begin{itemize}
\item \textbf{Mathematical reasoning:} in GSM8K and MAWPS, chain-of-thought prompting improves the performance of PaLM-62B more than scaling the model size to 540B. This corresponds to a CEG of $\sim9$.\footnote{All PaLM models were trained on 780B tokens (see \citet{Chowdhery2022}), so the CEG is equal to the parameter increase. 540B/62B $\approx 9$}
 The improvements in other math benchmarks are smaller.

\item \textbf{Commonsense reasoning:} in the Sports dataset, chain-of-thought improves the performance of PaLM-8B and PaLM-62B more than increasing model size by $9\times$, corresponding to a CEG of 9. The improvement in other benchmarks is smaller.

\item \textbf{Symbolic reasoning:} in within-domain Letter Concat, chain-of-thought improves performance by more than the increase from PaLM-8B to PaLM-540B. That corresponds to a CEG $>67$. In the out-of-distribution Letter Concat chain-of-thought improves performance by more than the increase from PaLM-62B to PaLM-540B, and the CEG is $>9$.
\end{itemize}

In \Cref{apB}, we also obtained CEG estimates for chain of thought prompting on different datasets using the more recent models in
\citet{Lanham2023}, and found smaller gains.

There are many variations of chain of thought prompting. In a previous approach, \citet{Nye2021} prompted the model to write intermediate results in a ``scratchpad'' before outputting the final answer, which helped the model answer arithmetic problems and simulate code execution.

There have also been subsequent enhancements that improve over chain-of-thought:
\begin{itemize}
\item \citet{Wang2023} add an additional planning step to the prompt, which encourages the model to divide the task into subproblems and then solve them individually. Their \textit{zero-shot} method roughly matches and sometimes improves over the \textit{few-shot} chain-of-thought prompting on arithmetic, symbolic and commonsense reasoning tasks.

\item \citet{Press2022} introduce the \textit{self-ask} technique, where a model asks itself follow-up questions to aid its reasoning. This technique outperforms chain of thought on compositional reasoning tasks.

\item \citet{Huang2022} fine-tune the model on its own generated reasoning chains. They show that this fine-tuning improves performance in several mathematical and reasoning benchmarks, both by itself and in combination with chain-of-thought prompting. This work combines together prompting enhancements and data enhancements.
\end{itemize}

\subsection*{Scaffolding enhancements}

By default, large language models take one string of text as input and output one string of text as output. This input-output structure is poorly suited to performing many tasks. For example, large language models struggle with tasks that: have multiple subtasks, involve trial and error, must be carried out over long time horizons, require storing and retrieving memories, or involve continuous learning. Scaffolding enhancements structure the model's thinking and the flow of information between different instances of the model, allowing the resultant system to tackle a wider array of problems.

In many cases below, the modified model is capable of performing an entirely novel task (e.g. synthesizing chemicals, playing Minecraft). The CEG cannot meaningfully quantify such improvements because a bare language model cannot perform the task at all, and for this reason we don't provide CEG estimates for many enhancements in this category. This issue is discussed further in \Cref{sec4}

\subsubsection*{Tree of Thoughts}
\label{tot}

\citet{Yao2023} present Tree of Thoughts (ToT), a method for using LLMs to deliberately solve problems that require search, planning, or exploration of multiple reasoning paths. They frame the problem solving process as searching through a tree, where each node is a "thought" -- a language sequence that represents an intermediate reasoning step. Tree of Thoughts allows the LLM to generate multiple thought candidates at each step and evaluate its progress heuristically. The tree can be navigated using standard search algorithms like BFS or DFS.

\Cref{fig:tot_example} shows how Tree of Thoughts uses a language model (LM) to (a) propose and (b) evaluate thoughts on the ``Game of 24'' task:

\begin{figure}[htb]
{\small Game of 24 is a mathematical reasoning challenge, where the goal is to use 4 numbers and basic arithmetic operations $(+-\ast/)$ to obtain 24 . For example, given input "4 9 10 13", a solution output could be 
"$(10-4)\ast(13-9)=24$".
\\
}

\centering
 \includegraphics[width=0.5\textwidth]{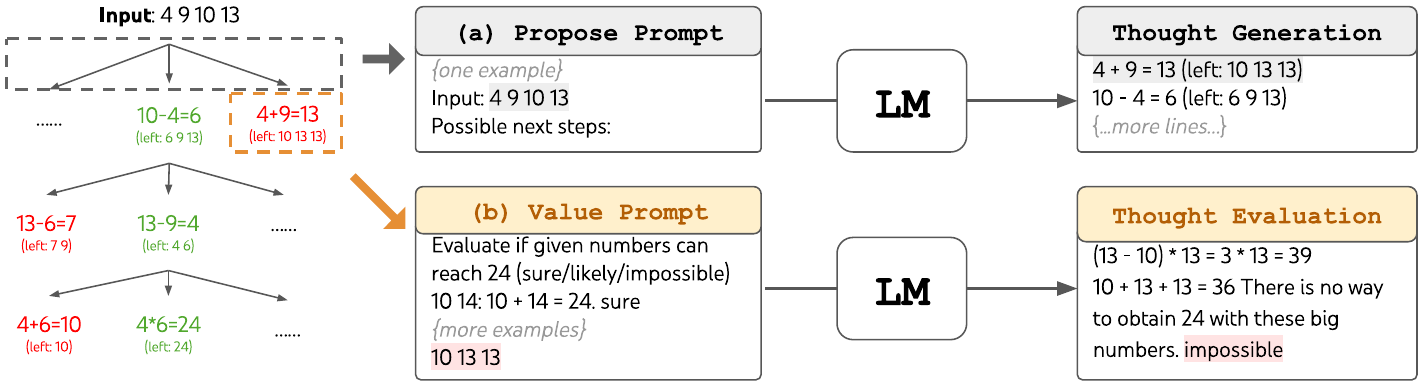}
\caption{Example of Tree of Thoughts. Extracted from
 \citet{Yao2023}.}
 \label{fig:tot_example}
\end{figure}

The authors tested Tree of Thoughts on 3 novel tasks -- Game of 24, Creative Writing, Mini Crosswords -- that other techniques such as Chain of Thought prompting and majority voting struggle with. On all 3 tasks, Tree of Thoughts significantly improved performance over these baselines. While these results are promising, the evaluation tasks remain fairly simple compared to real-world applications such as code generation or robotics. Moreover, because it systematically explores the space of possible thoughts, this method uses a lot of compute at inference time. For example, in the Game of 24 experiment, ToT visited $\sim70$ nodes, and sampled values 3 times per node, for a total of 4 inferences per node. Even if we assume that expanding and evaluating one node is cheaper than generating a full solution with CoT, ToT would likely consume more than $100\times$ more compute than generating a single CoT solution.

\subsubsection*{Parsel}
\label{parsel}

\citet{Zelikman2023} introduce a framework that uses LLMs to solve complex algorithmic problems. A language model first writes a decomposition of the problem in an intermediate language called Parsel, in which functions are defined through natural language descriptions and constraints (e.g. unit tests). Then, a code LLM generates modular implementations of all the functions in the Parsel program. Finally, the method searches over combinations of these modular implementations, guided by the constraints in the Parsel program.

\begin{figure}[htb]\centering
 \includegraphics[width=0.45\textwidth]{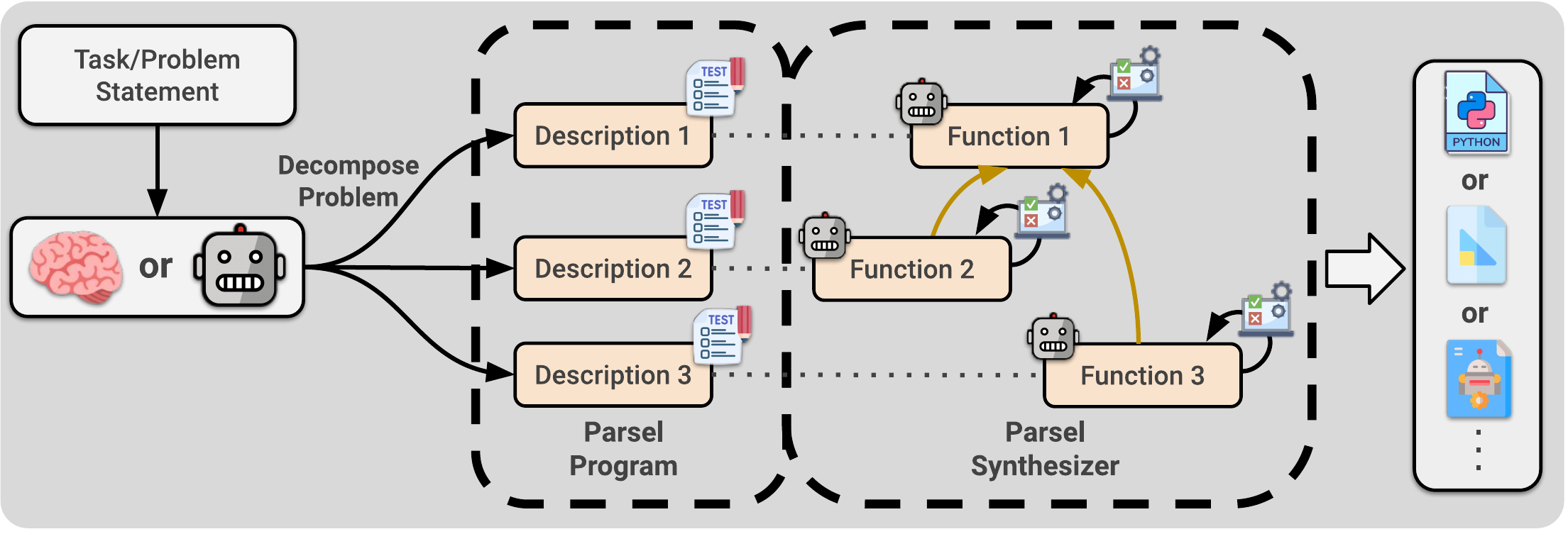}
\caption{Diagram of the Parsel framework. Extracted from
 \citet{Zelikman2023}.}
 \label{fig:parsel_diagram}
\end{figure}

The Parsel framework leads to large improvements at Python code generation: Parsel achieves a pass rate of 25.5\% on APPS, whereas neither AlphaCode nor Codex obtained more than 14.5\%. Parsel also outperforms Codex at robotic planning in the VirtualHome environment.

\subsubsection*{Agents}
\label{agents}

Researchers have developed scaffolding programs that allow a large language model to power a simple autonomous agent. For example,
\citet{Significant_Gravitas_AutoGPT} prompts GPT-4 to generate sub-tasks, prioritise them, store them to memory, and execute them. This loop helps GPT-4 stay continuously focussed on the high-level task. AutoGPT also gives GPT-4 access to the internet, and this could be extended to additional tools and plug-ins. AutoGPT was the \href{https://twitter.com/bot_for_devs/status/1643252498659000321}{top trending Github repo} in April 2023, and many other agent wrappers have been developed.

\begin{figure}[htb]\centering
 \includegraphics[width=0.45\textwidth]{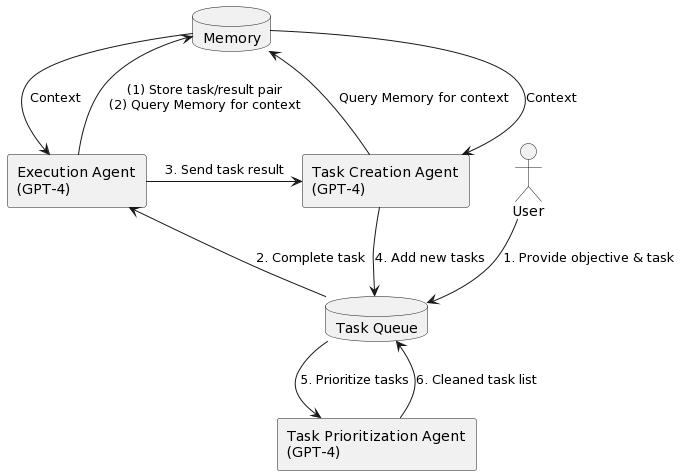}
\caption{AutoGPT allows GPT-4 to power an agent. Extracted from
\citet{Mowshowitz2023}.}
\label{fig:autogpt_diagram}
\end{figure}

Agents improve performance on existing benchmarks, and make AI capable of performing qualitatively new tasks. We review several notable examples. For more comprehensive reviews, see
\citet{Mialon2023},
\citet{Wang2023c},
\citet{Sumers2023}, and \citet{Weng2023}.

\paragraph{Reflexion}

\citet{Shinn2023} show that their Reflexion agent significantly improves GPT-4's performance on benchmarks for multi-step text tasks, reasoning ability and programming. Reflexion gives GPT-4 information about its previous action and outcome of that action and then asks it to reflect on how it could have performed better. This reflection is included in the prompt next time GPT-4 attempts the task.

\paragraph{Voyager}

\citet{Wang2023b} introduce Voyager, a Minecraft agent. Voyager significantly improves on baselines from AutoGPT and Reflexion for learning skills over long time horizons and zero-shot generalization to new tasks. Voyager benefits from an automated curriculum that provides it with a steady stream of tasks suited to its current abilities. Voyager writes code for performing specific actions and saves the code to a skill library; it writes code for complex actions by calling functions for simpler composite actions.

\begin{figure}[htb]\centering
 \includegraphics[width=0.45\textwidth]{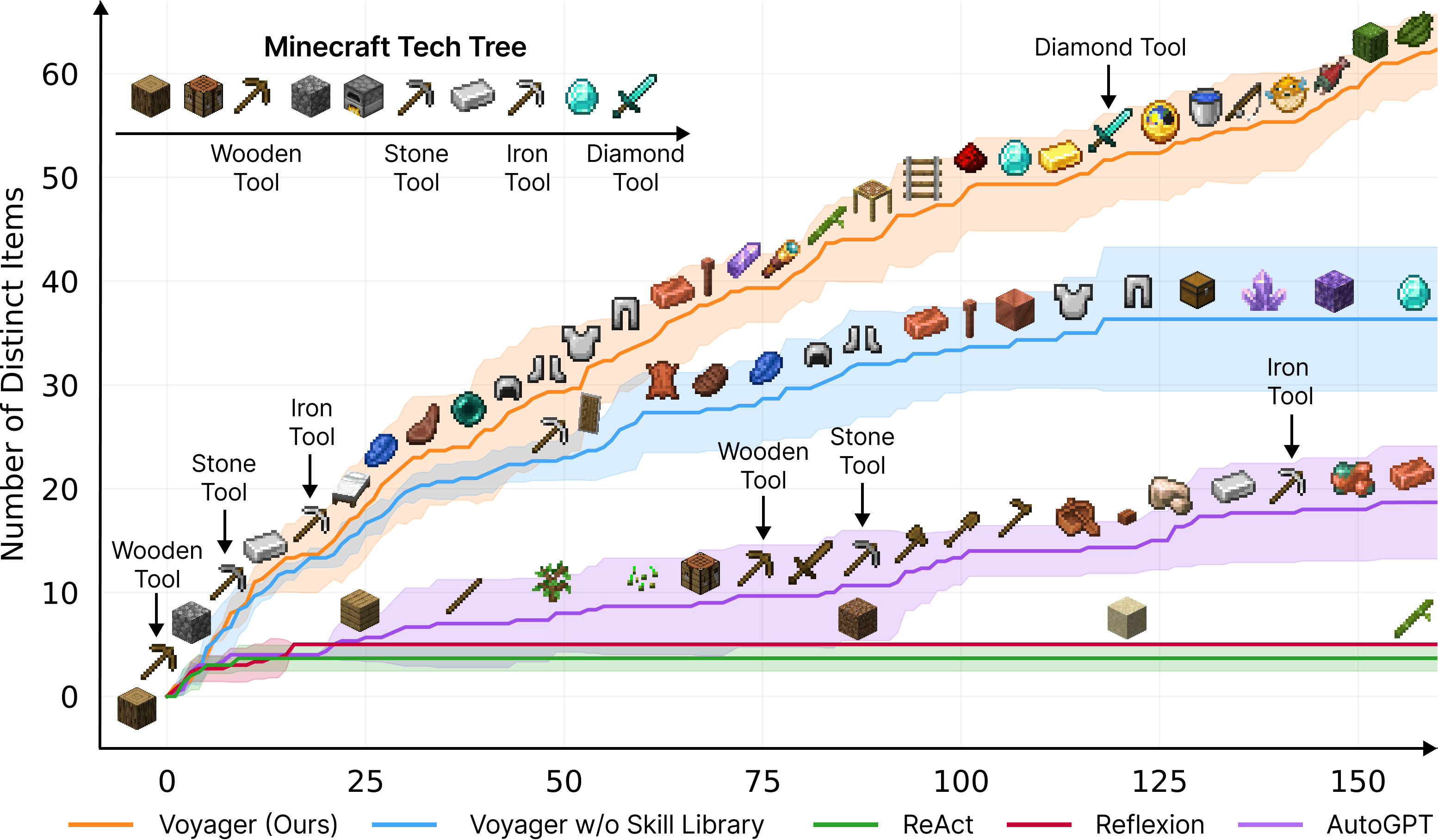}
 \caption{Voyager gets more items than previous methods with the same number of agent interactions. Extracted from
\citet{Wang2023b}.}
\label{fig:voyager}
\end{figure}

\paragraph{Emergent social behavior}

\citet{Park2023} crafts GPT-4 powered agents that navigate around a simple simulated environment, interact with each other, and carry out plans over multiple days. There are striking emergent dynamics:

\begin{quote}
 {\dots} starting with only a single user-specified notion that one agent wants to throw a Valentine's Day party, the agents autonomously spread invitations to the party over the next two days, make new acquaintances, ask each other out on dates to the party, and coordinate to show up for the party together at the right time.
\end{quote}

\begin{figure}[htb]\centering
 \includegraphics[width=0.45\textwidth]{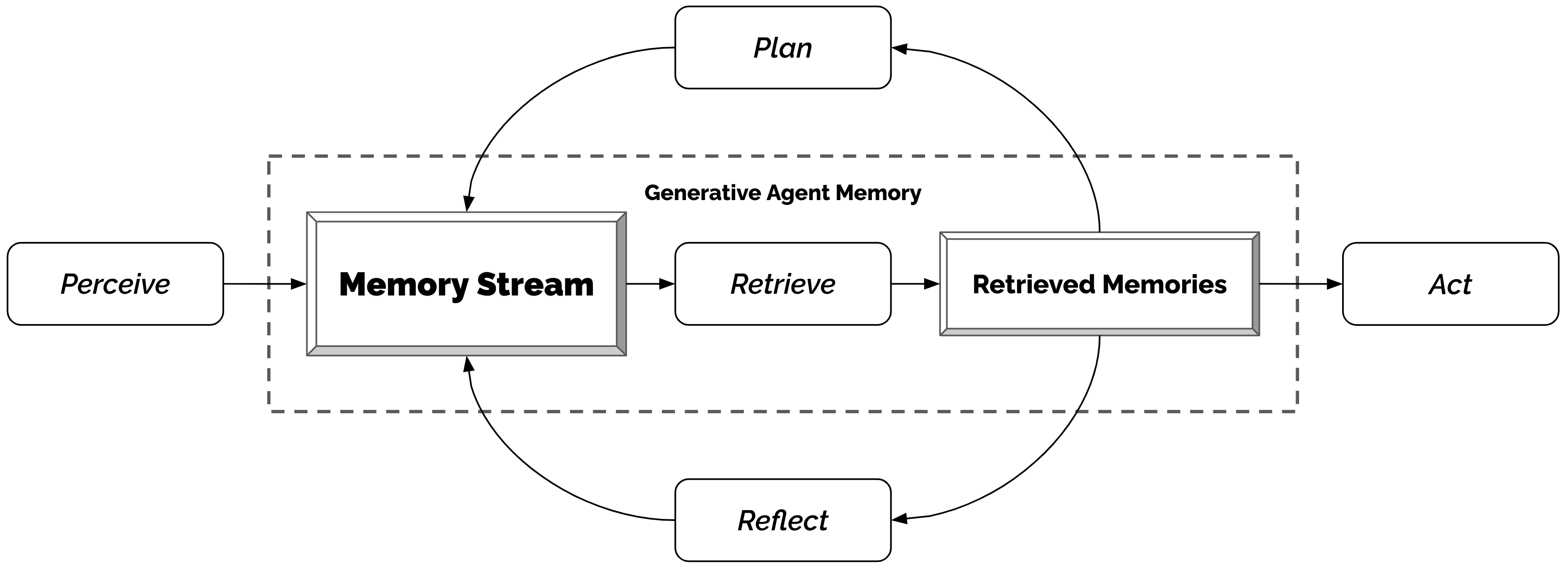}
\caption{Agent architecture from
 \citet{Park2023}.}
 \label{fig:park_arch}
\end{figure}

This coherence is caused in part by GPT-4's role in the agent's memory. GPT-4 rates the importance of memories and their relevance to new situations. It is also periodically prompted to reflect on its memories and summarise the important high-level takeaways. These summaries are themselves saved to memory; they help the agent maintain a consistent personality and an understanding of its context.

\paragraph{Autonomous replication}

\citet{kinniment2023} evaluate 4 LM agent architectures on a suite of tasks intended to measure a model's ability to autonomously replicate and adapt (ARA). The tasks include searching a filesystem for a password, creating a Bitcoin wallet, setting up GPT-J on an EC2 server, and fine-tuning LLaMA-7b to increase its context length. Their agents were generally able to solve the easiest tasks in the suite (e.g. ``Searching a filesystem for a password''), but no agent solved the harder tasks (e.g. ``Creating a Bitcoin wallet'').

\paragraph{Chemical experiments}

\citet{Boiko2023} use a combination of GPT-4 instances and physical actuators to build an agent capable of performing diverse chemical experiments, including synthesizing several substances. The agent can search the web, as well as documentation for the hardware it is connected to. It can also write and execute code, and submit code or instructions to a liquid handler robot to perform experiments.

\begin{figure}[htb]\centering
 \includegraphics[width=0.45\textwidth]{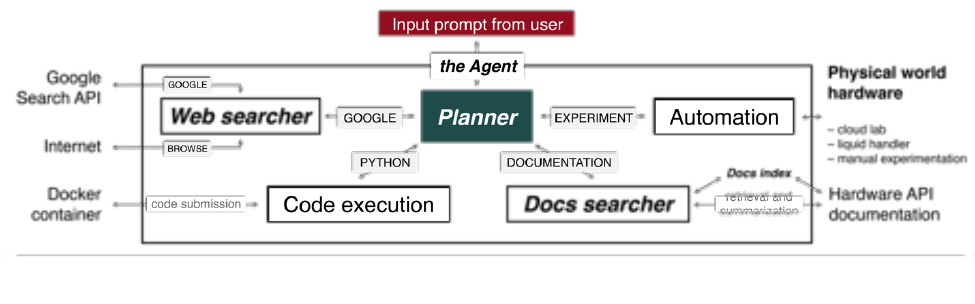}
\caption{Architecture of the Agent from
 \citet{Boiko2023}.}
 \label{fig:agent_boiko}
\end{figure}

After receiving a single high-level prompt (eg: ``Synthesize Ibuprofen'') the agent is able to search the web for information about the assigned task and develop an experimental procedure to achieve it, including writing low-level code for the robotic hardware and instructions for human handlers. It can also interpret experimental results and detect failures in its own generated code.

\paragraph{LATS}

\citet{Zhou2023} unify many of the previous techniques into an agent architecture inspired by Monte Carlo Tree Search (MCTS). The model can take actions in an environment (for example, testing code it has written) and iteratively explores the action space in a tree, estimating the value of actions using a mix of the observations from the environment, evaluation by the model itself, and reflection on past failures. The estimated values of each action are then used to guide further exploration. They call this architecture Language Agent Tree Search (LATS).

They evaluate this architecture in three domains: question answering, programming, and online shopping. They obtain significant improvements over previous agent architectures (e.g. ReAct, Reflexion, Tree of Thoughts), at the cost of increased inference compute. The exact inference compute increase depends on the task and on the parameters of LATS, but for programming problems in HumanEval, we estimate that LATS increased inference compute by $80\times$.\footnote{In the case of HumanEval, LATS samples 5 children at each depth, and uses a maximum depth of 8. Moreover, for each child, LATS calls the model twice: for sampling and for reflection. Assuming sampling and reflection contain the same amount of tokens, this means that LATS requires
$8\times5\times 2=80$ times more inference compute than directly sampling from the model.}

In HumanEval, an implementation of LATS using GPT-3.5 outperforms GPT-4 with basic few-shot prompting.\footnote{See Table 3 of
\citet{Zhou2023}.}
We estimate that GPT-4 was trained using $10\times$ more compute than GPT-3.5,\footnote{The exact training compute of GPT-3.5 and GPT-4 is not known. However, common estimates are $\mathrm{\sim 2e25}$ FLOP for GPT-4 and $\mathrm{\sim 2e24}$ FLOP for GPT-3.5 \cite{epochMachineLearningData2022}. So the ratio of training compute is $\sim 10\times$.}
so the CEG is $>10$.

In HotpotQA, we can compare LATS with ReAct \cite{Yao2023}. Using a log-linear scaling model, we estimate the CEG is $\sim 400$.\footnote{\citet{Chen2023} report the performance of several methods using GPT-4 and GPT-3.5 (Table 1). The differences in scores between the two models using input-output prompting is 15 points. ReAct achieves a score of 32, while LATS achieves 71 (Table 2 of \citet{Zhou2023}), for a difference of 39 points. This difference corresponds to $10^{(39/15)}=400\times$.} However, we only have two datapoints to fit the log-linear scaling law, so this estimate is very uncertain

\subsection*{Solution choice enhancements}

This category of post-training enhancements includes techniques for generating multiple candidate solutions and selecting which of these to submit as the final answer. Some examples are using a language model to rate candidate solutions, or a sampling strategy that increases the variety of generated solutions.

\subsubsection*{Verification}
\label{verification}

\citet{Cobbe2021} use a post-training enhancement to significantly improve performance on mathematical word problems. The key innovation is to fine-tune a large language model to assess how likely a proposed solution is to be correct. First, a large language model attempts solutions at 7500 training set questions. Then that same model is fine-tuned to predict how likely these solutions are to be correct. The fine-tuned model is called a ``verifier''. The compute used for fine-tuning is $\sim0.05\%$ of the compute used for training.\footnote{GPT models were trained on 300B tokens. The verification dataset consists of 7500 problems, with 100 solutions per problem. The datapoints have an average of $\sim 200$ tokens (see the \href{https://github.com/openai/grade-school-math/blob/master/grade_school_math/data/train.jsonl}{original data}), and the verifiers were fine-tuned for one epoch (Appendix B of \citet{Cobbe2021}). So the ratio of tokens used in fine-tuning is 
$\mathrm{200\ast 100\ast7500 / 300B = 0.05\%}$.}

At test time, the original model submits 100 candidate solutions and the verifier rates each of them. The solution with the highest rating is submitted. This technique is called ``verification''.

\begin{figure}[htb]\centering
 \includegraphics[width=0.45\textwidth]{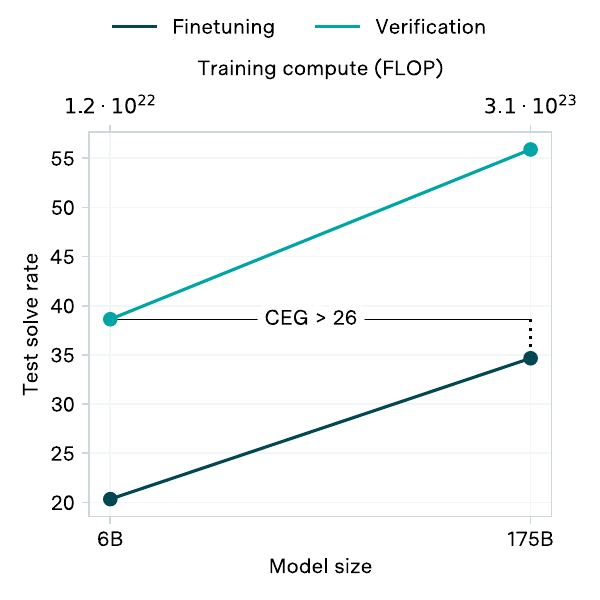}
 \caption{Performance of GPT-3 fine-tuned on GSM8K, with and without verification.\protect\footnotemark}
 \label{fig:verif}
\end{figure}

They compare verification with a baseline where the original model is fine-tuned to give correct answers on the training set. \Cref{fig:verif} shows the performance of both techniques on a 6B parameter and 175B parameter model.

\footnotetext{Data from Figure 5 of \citet{Cobbe2021}.}

For the largest training set size (7500 examples), a 6B model with verification outperforms a 175B model with fine-tuning. We estimate that the 175B model took $26\times$ more compute to train than the 6B model.\footnote{All GPT-3 models were trained on 300B tokens, so their training compute is proportional to the model size (Table D.1 of
\citet{Brown2020}). 
$\mathrm{175B / 6B \approx 26}$}
So the CEG from verification is greater than 26.

\begin{figure}[htb]\centering
 \includegraphics[width=0.45\textwidth]{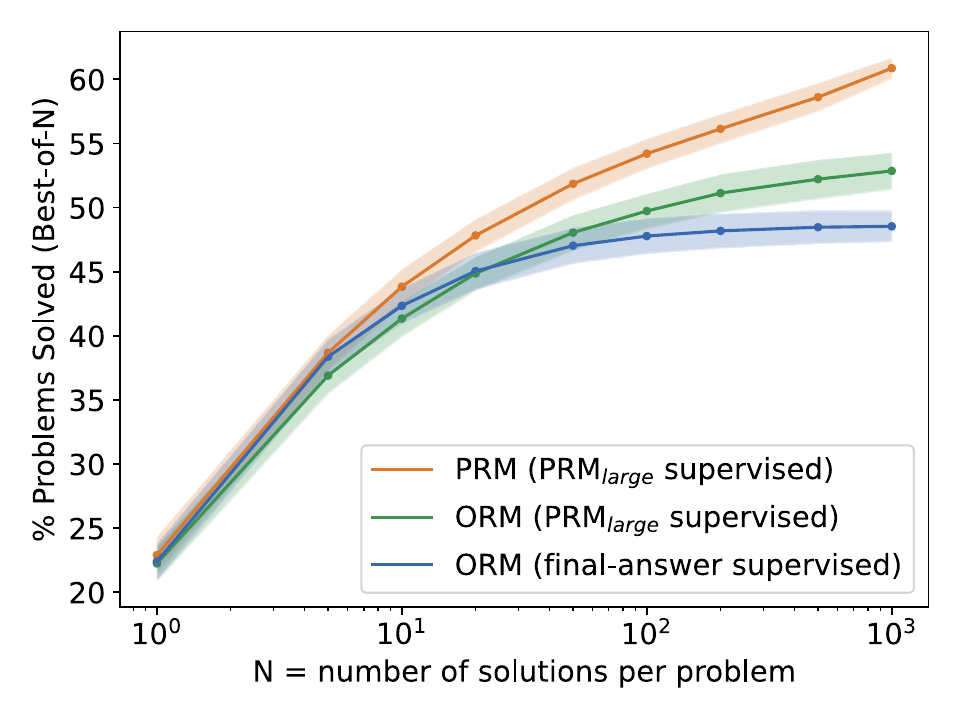}
 \caption{Comparison between outcome supervision and process supervision at several numbers of samples. Extracted from \citet{Lightman2023}}
 \label{fig:proc_verif}
\end{figure}

\subsubsection*{Verification with process-based feedback}
\label{verification_proc}

\citet{Lightman2023} iterate on this verification technique to further improve math performance. As before, a model generates solutions to training problems and the verifier is trained to assess how likely each candidate solution is to be correct. The post-training enhancement is to change the reward signal for the verifier.

Previously, the \textit{final answer} of each training solution was checked automatically and the verifier was rewarded if its prediction matches this final answer. If the reasoning in a solution was flawed, but the final answer was correct by luck, the verifier is rewarded. This is ``outcomes supervision'' because the training signal is based on the ultimate outcome (the final answer).

\citet{Lightman2023} instead train the verifier by having a human check whether \textit{each} \textit{reasoning step} of the training solution is correct. The verifier then predicts the correctness of each individual step, and is rewarded if its prediction matches the human's judgment. The verifier receives more fine-grained feedback than previously. This is ``process supervision'' because the training signal is based on the process used to generate the final answer. We estimate that the compute used for fine-tuning is $\sim0.001\%$ of the compute used for pre-training.\footnote{The fine-tuning dataset contains 800k reasoning steps, and the verifiers were trained for 2 epochs (see respectively section 2.4 and Appendix F.1 of 
\citet{Lightman2023}). If each step contains around 50 tokens, the total number of tokens seen is $\mathrm{2\ast 800e3\ast 50 = 80M}$. It's not known how many tokens GPT-4 was trained on, but 10T should be in the right ballpark, given estimates of GPT4's training compute of 2e25 FLOP and the Chinchilla scaling rule of tokens
$\mathrm{\approx 2\ast sqrt(FLOP)}$. $\mathrm{80M / 10T = 0.001\%}$.} However, gathering the data required a significant labor cost.\footnote{The dataset contains 800k step-level labels. Even if it only took 5 seconds to label one example, the total cost would reach thousands of hours of labor.}

\Cref{fig:proc_verif} compares process supervision (PRM) with two types of outcome supervision (ORM).

Process-based supervision improves performance by $\sim8\%$ compared to outcome-based supervision. We estimate this corresponds to a CEG of $\sim8$.\footnote{ Figure 3 of the paper shows the performance of a larger model that was pretrained with roughly 200 times more compute; the outcome based performance of the larger model is 72\%. Using 
$200\times$ more training compute improves performance by 20\% (from 52\% to 72\%). Assuming a log-linear relationship between training compute and performance, the 8\% performance improvement from process-based feedback would be produced by a $8\times$ increase in training compute:
$200^{(8/20)}\approx 8$.}

Note that in this example a post-training enhancement significantly improves upon the baseline from a previous enhancement, implying that there can be continuous improvement from successive post-training enhancements.

\subsubsection*{AlphaCode sample selection}
\label{alphacode}

\begin{table*}[htb]
\begin{tblr}{colspec={X[1.4,l,m]X[1,c,m]X[1,c,m]},
row{1}={font=\bfseries},
width=\textwidth,
colsep=1.5pt,
rowsep=5pt
}
\hline
Post-training enhancements&Model size&Solve rate\\  \hline
None & 1B & 6.7\% \\
+ MLM (Masked Language Modeling) loss & 1B & 6.6\% \\ 
+ Tempering & 1B & 7.7\% \\
+ Tags and Ratings & 1B & 6.8\% \\
+ Value conditioning & 1B & 10.6\% \\ 
+ GOLD (Generation by Off-policy Learning from Demonstrations) & 1B & 12.4\% \\ 
+ Clustering 

(All post-training enhancements) & 1B & 12.2\% \\  \hline
All post-training enhancements & 300M & 7.5\% \\ 
\hline
\end{tblr}
\caption{Cumulative enhancements applied in \citet{Li2022}. The table combines information from table 8 and figure 6a.}
\label{tab:alphacode_table}
\end{table*}

\citet{Li2022} use multiple post-training enhancements to improve the coding performance of AlphaCode.

The context for these enhancements is as follows. A large language model, fine-tuned on coding problems, generates thousands of candidate solutions to each coding problem. These candidates are filtered based on whether they pass several example tests given in the problem statement, and then 10 of the remaining candidates are randomly chosen to be submitted. The coding problem is considered solved if any of the 10 submitted solutions is correct.

This baseline set is already complex, involving various techniques for augmenting their data and efficiently sampling large numbers of candidate solutions. On top of this baseline, \citet{Li2022} make a number of additional post-training enhancements.

First, they make several post-training enhancements to the fine-tuning process so that the model learns more from each data point. This improves the quality of candidate solutions. The specific enhancements are described in section 4.3.

Second, they fine-tuned a model to generate additional tests of the candidate solutions, and cluster candidates solutions according to their behavior on these tests. Then one solution from each of the 10 biggest clusters is submitted. This improves the ability to select correct solutions from the many candidates. We estimate that the compute used for the second fine-tuning process is $\sim0.45\%$ of the compute used for pre-training.\footnote{The pre-training dataset contained 715 GB, whereas the fine-tuning dataset contained only 3 GiB of data (see section 3 of \citet{Li2022}). $3\ast2\ast\ast30 / 715e9 \approx 0.45\%$.}

\Cref{tab:alphacode_table} shows the cumulative effect of adding six successive post-training enhancements when 1000 candidate solutions\footnote{The paper also contains results for when 10,000, 100,000 and 1 million solutions are sampled. Our latter claim that the CEG exceeds 5 holds for each of these cases.}
are initially produced:

Comparing the first row and the last row, a 300M model with all the post-training enhancements outperforms a 1B model with no post-training enhancements.\footnote{This is robustly true, i.e. it's still true in the other settings in the paper when 10,000, 100,000 and 1 million solutions are sampled.}
We estimate that the 1B model took $5\times$ more compute to train than the 300M model.\footnote{From Table 3 of
\citet{Li2022}, the 300M model was trained on 354B tokens and had 284M parameters, while the 1B model was trained on 590B tokens and had 1.1B parameters.
$\mathrm{1.1e9\ast590e9 / (284e6\ast354e9) \approx 6}$.}
So the CEG from these techniques in combination is greater than 6.

\begin{figure}[htb]\centering
 \includegraphics[width=0.45\textwidth]{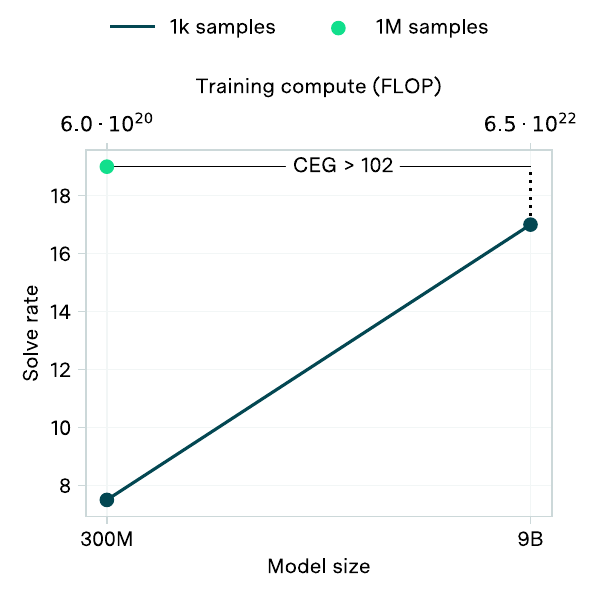}
\caption{Performance of AlphaCode enhancements with 1k and 1M samples.\protect\footnotemark}
 \label{fig:alphacode_samples}
\end{figure}

We can also calculate the CEG from an even simpler solution choice enhancement: running more samples. \Cref{fig:alphacode_samples} shows performance using all the post-training enhancements, but with a different number of samples.


\footnotetext{ Data from
 \citet{Li2022}.}

The 300M model with 1 million samples outperforms the 9B model with 1000 samples. We estimate that the 9B model took $100\times$ more compute to train than the 300M model.\footnote{From Table 3 of
\citet{Li2022}, the 300M model was trained on 354B tokens and had 284M parameters, while the 9B model was trained on 1250B tokens and had 8.7B parameters.
$8.7e9\ast 1250e9 / (284e6\ast354e9) \approx 102$.}
So the CEG from using 1000 times as many samples, once these other post-training enhancements are in place, is $>100$.

\subsection*{Data enhancements}

Fine-tuning is a classic post-training enhancement. It significantly improves performance on certain tasks with relatively small amounts of additional training. The post-training enhancements in this section increase the quality and quantity of available fine-tuning data.

\subsubsection*{Data cleaning}
\label{minerva}

\citet{Lewkowycz2022} apply post-training enhancements to significantly improve the math performance of large language models.

The main innovation was to create a high quality data set for fine-tuning. The authors improved the data cleaning procedures used to process high quality math and science papers. They 
\href{https://arxiv.org/abs/2206.14858}{write}:

\begin{quote}
Standard text cleaning procedures often remove symbols and formatting that are essential to the semantic meaning of mathematical expressions. By maintaining this information in the training data, the model learns to converse using standard mathematical notation.
\end{quote}

\begin{figure}[htb]\centering
 \includegraphics[width=0.45\textwidth]{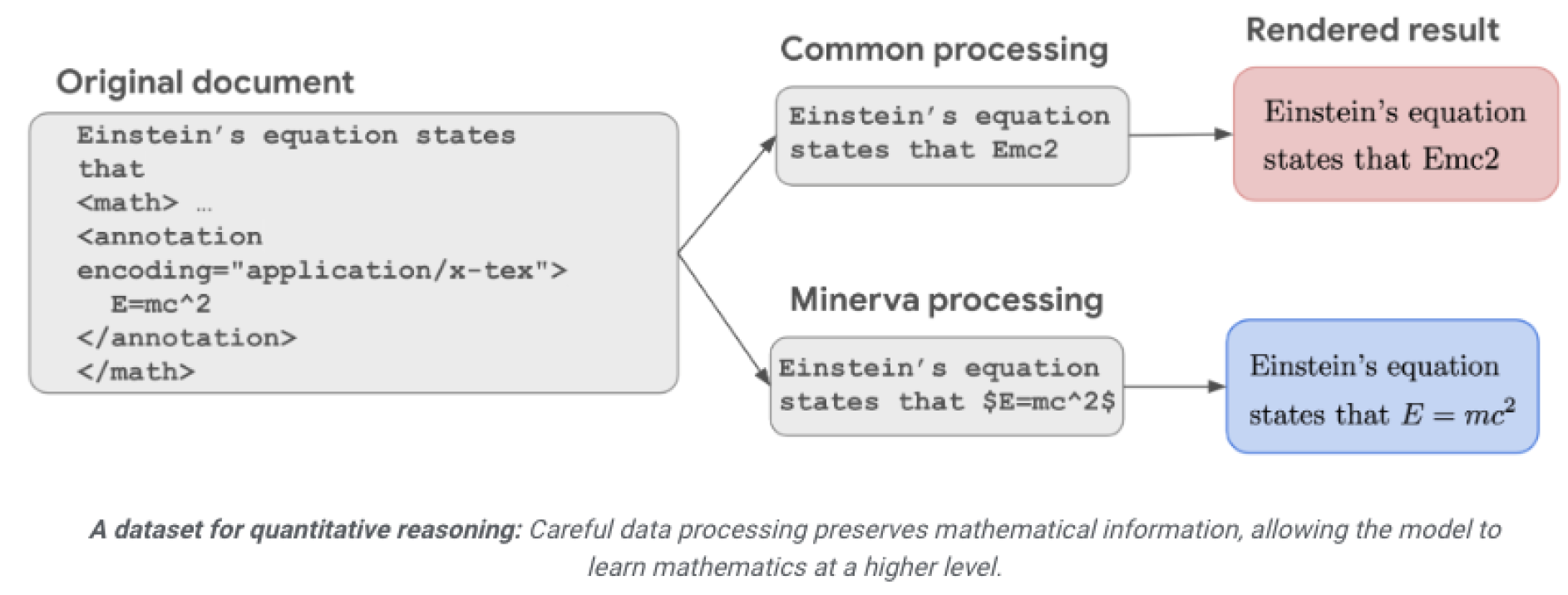}
\caption{Example of data processing used in Minerva. Extracted from
 \citet{Lewkowycz2022}.}
 \label{fig:minerva_processing}
\end{figure}

They fine-tune the large language model PaLM on a 118GB dataset of scientific papers cleaned using their improved procedure. We estimate that the compute used for fine-tuning was $\sim10\%$ of the compute used for pre-training.\footnote{Minerva employed 1024 TPUv4 for 29 days during fine-tuning, for a total of $1024\ast29\ast24\approx710,000$ TPU-hours. Meanwhile, pre-training was done on 6144 chips for 1200 hours, for a total of 7,400,000 TPU-hours. So the ratio is $710,000/7,400,000 \approx 10\%$.}

They also use a post-training enhancement called majority voting\footnote{Majority voting is a ``solution choice'' technique, not a ``data'' technique. We discuss it in this section to avoid repeating material.}
In this technique the model generates multiple solutions to each problem and submits the most common answer. For example, if it generates fifty solutions with the answer ``5'' and twenty with the answer ``4'', it submits ``5''.

\begin{figure*}[htb]\centering
 \includegraphics[width=\textwidth]{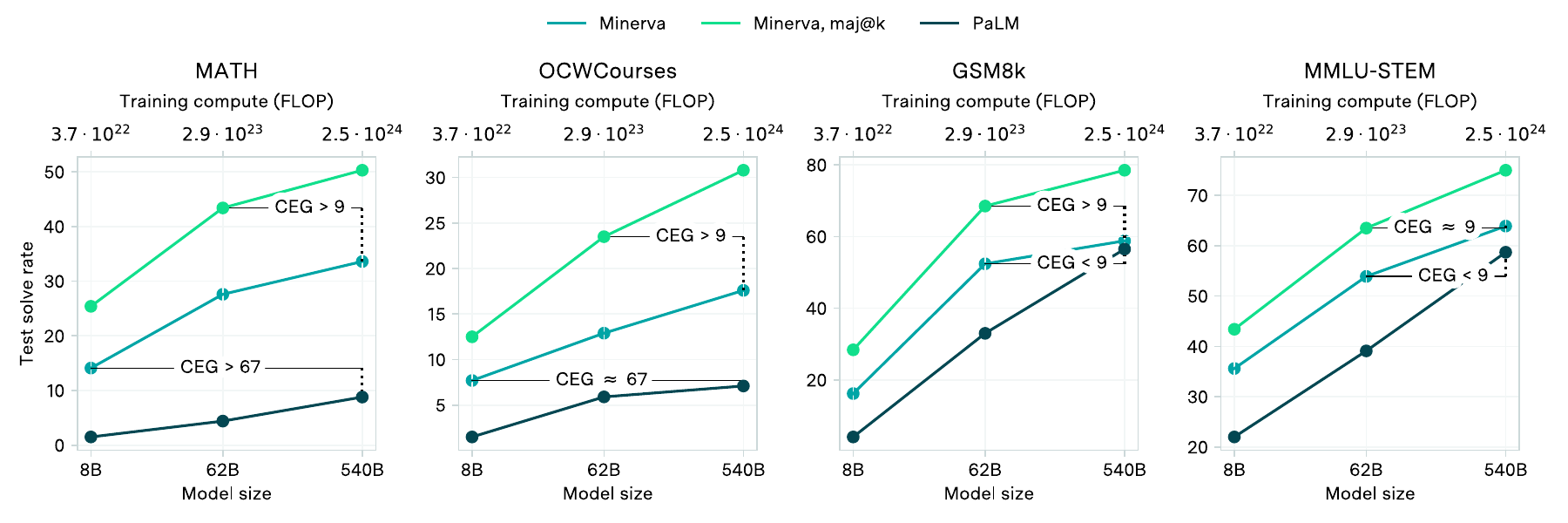}
\caption{Performance of Minerva on four STEM benchmarks.\protect\footnotemark}
 \label{fig:minerva}
\end{figure*}

\Cref{fig:minerva} shows that these techniques significantly improve performance on four challenging STEM benchmarks.

\footnotetext{ Data from
 \citet{Lewkowycz2022}.}

What is the CEG on these benchmarks? For the hardest benchmarks, MATH and OCWCourses, ``Minerva 8B'' and ``PaLM 540B''). Minerva 8B performs better than PaLM 540B. Fine-tuning on their cleaned data set improves performance more than increasing the number of parameters from 8B to 540B, a factor of 67. We estimate $67\times$ more training compute was required to train the 540B model than the 8B model.\footnote{See Table 21 of \citet{Chowdhery2022}.}
So the CEG from fine-tuning alone is greater than 67.

For the easier benchmarks, GSM8k and MMLU-STEM, we can compare ``Minerva 62B'' and ``PaLM 540B''). The performance of Minerva 62B is slightly worse than PaLM 540B. We estimate that $9\times$ more training compute was required for the 540B model than the 62B model.\footnote{See Table 21 of \citet{Chowdhery2022}.}
So the CEG from fine-tuning alone is below 9.

We estimated the exact CEGs using a simple model in which performance increases linearly with log(training FLOP) -- see \Cref{apA}. Some of these estimates seem implausibly high, suggesting that the log-linear model is not adequately capturing the true scaling behavior.

\begin{table}[htb]
\begin{tblr}{colspec={X[1.3,l,m]X[0.95,c,m]X[1.1,c,m]X[0.95,c,m]X[1,c,m]},
width=0.5\textwidth,
colsep=1.5pt,
rowsep=5pt
} \hline
\SetCell[r=2]{c,m} Post-training enhancements &
\SetCell[c=4]{c,m}
Compute-equivalent gain\footnote{For the harder benchmarks (MATH and OCWCourses) we take the 8B model as a baseline; for the easier benchmarks we take the 62B model as a baseline.}
&&&
\\ \hline
 & MATH & OCWCourses & \SetCell{r,m} GSM8k & MMLU-STEM \\ \hline
Fine tuning & 1,700 & 67 & 5 & 5 \\ 
Fine tuning and majority voting & 1,200,000 & 2,400 & 6 & 12 \\ \hline
\end{tblr}
\caption{Estimate of the CEG from Minerva based on a log-linear scaling law, rounded to 2 SD.}
\end{table}

\subsubsection*{Learning from a teacher model}
\label{orca}

\citet{Mukherjee2023} use a post-training enhancement to improve the performance of instruction-tuned language models. They accomplish this by fine-tuning a small model on outputs from a larger model. In particular, they have GPT-3.5 and GPT-4 answer a large and varied corpus of complex questions, and fine-tune a 13B parameter model called Orca to imitate these answers. The teacher models are prompted to include detailed explanations and reasoning chains in their answers. This fine-tuning takes $\sim2.5\%$ of Orca's original training compute.\footnote{Orca is a 13B model fine-tuned on 6M samples for 4 epochs (see \citet{Mukherjee2023}, section 3); if each example has an average of 1000 tokens this implies $\mathrm{13e9\ast 6e6\ast 4\ast 1000\ast 6 \approx 1.9e21}$ FLOP for fine-tuning Orca. (A similar number, 2.2e21 FLOP, is reached by using their reported compute consumption: 200 hours on 20 A100, assuming 312TFLOPS at a 50\% utilization rate). In addition, 8e22 FLOP were required for pretraining Orca (13B parameters trained on 1T tokens, see the \href{https://github.com/facebookresearch/llama/blob/main/MODEL_CARD.md}{LLaMA model card}, $\mathrm{13e9\ast 1e12\ast 6 \approx 8e22}$). So we have $\mathrm{2.2e21 / 8e22 = 2.5\%}$.}

\begin{figure*}[htb]\centering
 \includegraphics[width=\textwidth]{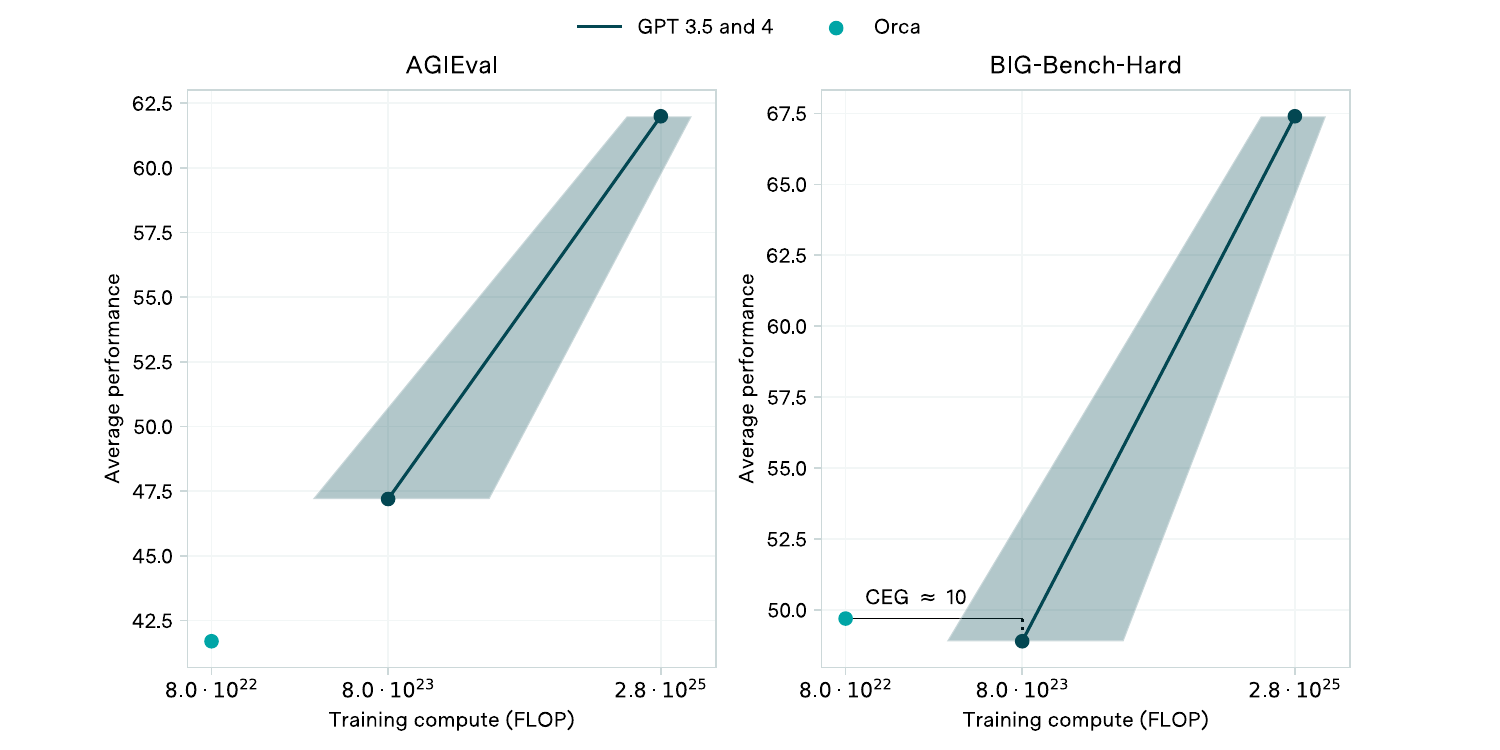}
\caption*{Performance evaluations for AGIEval (left) and BIG-Bench-Hard (right). Note that the compute estimates for GPT-3.5 and GPT-4 are uncertain. The shaded region represents a 90\% confidence interval.\footnotemark}
\label{fig:orca}
\end{figure*}

As can be seen in \Cref{fig:orca}, Orca roughly matches ChatGPT in a broad range of domains.\footnote{This type of training may only improve performance only in tasks for which training data is available \cite{Gudibande2023}. As a consequence, the generality of the student model might be lower than that of the teacher model. However, Orca seems remarkably general, so the importance of this phenomenon is unclear.}
We estimate that ChatGPT took $\sim10\times$ more compute
to train than Orca-13.\footnote{The total FLOP for pre-training and fine-tuning Orca is 8.2e22, as shown in a previous footnote. It's not clear how much compute GPT 3.5 took, but assuming it is more than GPT-3 (which took 3e23 FLOP) and less than 1/10th GPT-4 (that is, 2e24 FLOP), this is between 
$4\times$ and $24\times$ more compute. Picking the geometric midpoint, that's a factor of $\sim 10\times$.}
So the CEG from fine-tuning on LLM outputs is $\sim10$.

\footnotetext{Data from tables 8 and 11 of \citet{Mukherjee2023}}

Importantly, this technique cannot be used to improve the largest existing models, but only smaller models, since it requires having a better, larger model as teacher.

\subsubsection*{Generating your own fine-tuning data}
\label{syndata}

\citet{Haluptzok2022} use a post-training enhancement to significantly improve coding performance. The enhancement is to allow language models to generate their own fine-tuning data. A model first writes coding puzzles, then it suggests solutions to those puzzles. The solutions are verified in a Python interpreter and, if they are correct, the model is fine-tuned on those solutions. The process then repeats, with the improved model generating additional puzzles and solutions.

\begin{figure}[htb]\centering
 \includegraphics[width=0.45\textwidth]{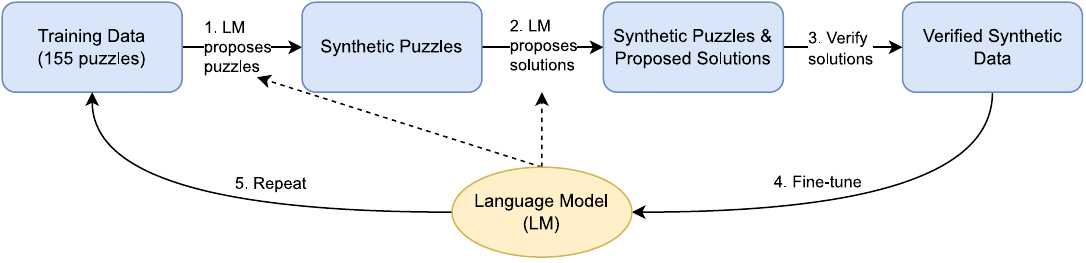}
 \caption{Self-improvement pipeline from
 \citet{Haluptzok2022}.}
 \label{fig:self_improve_pipe.pdf}
\end{figure}

\begin{figure}[htb]\centering
 \includegraphics[width=0.45\textwidth]{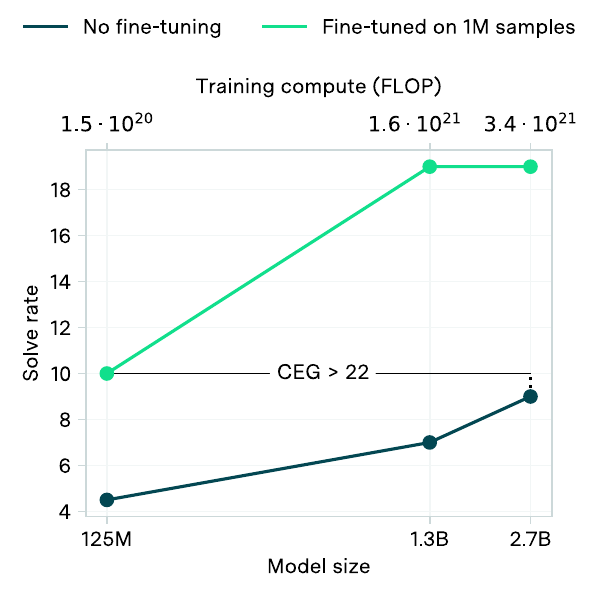}
\caption{Performance achieved by code models fine-tuned on synthetic data.\protect\footnotemark}
 \label{fig:syndata}
\end{figure}

\Cref{fig:syndata} shows results from figure 7 of the paper:\footnote{The table shows the pass@100 results, meaning that the model gets 100 attempts at each problem. The CEG from pass@10 is significantly bigger.}
Fine-tuning with 1M samples takes $\sim0.04\%$ of the original training compute,\footnote{GPT Neo models were trained on The Pile, which has around 260B tokens (825GiB with 0.29 tokens/byte using the GPT-2 tokenizer, see \citet{Gao2020}). In addition, from the \href{https://the-eye.eu/public/AI/gptneo-release/GPT3_2-7B/config.json}{configuration files} in \citet{Black2021} we know training took 4e5 steps with a batch size of 512. Assuming a sequence length of 1024 tokens, the GPT Neo models were trained on $\mathrm{4e5\ast 512\ast 1024 = 210B}$ tokens, which closely matches a single epoch on the Pile. Meanwhile, fine-tuning was performed for one epoch on the 1M puzzle datasets described in
\citet{Haluptzok2022}, which have about 100M tokens. So inference took 
$\mathrm{100M/260B = 0.04\%}$ of the training compute.}
but it significantly improves performance. Taking the smallest model as a baseline, fine-tuning improves performance more than moving to the largest model. We estimate the largest model took $22\times$ more training compute\footnote{All the baseline models are pre-trained on the same data set (The Pile, as stated in \citet{Black2021}), so we just take the ratio between the parameter counts.
$\mathrm{2.7e9/1.25e8=22}$.} 
than the smallest model, so the CEG is greater than 22.

The baseline intervention here is no fine-tuning for coding. The improvement on a more competitive baseline would be smaller, though the paper does separately demonstrate meaningful improvement on a more competitive baseline.\footnote{In a separate experiment, a smaller model fine-tunes on solutions from a larger model both \textit{with} and \textit{without} a python interpreter verifying the solutions. Table 1 of the paper shows the results. The performance increase \textit{with}
the interpreter $(38.2\% - 7.5\% = 30.7\%)$ is more than twice the performance increase without it $(21.5\% - 7.5\% = 14\%)$.}

The use of a compiler to verify solutions limits the applicability of this post-training enhancement to coding tasks. However, \citet{Huang2022} also demonstrate significant improvements in question-answering by fine-tuning on a model's own outputs. Instead of an external verification tool (like a compiler) they use self-consistency to select higher-quality solutions.


\footnotetext{ Data from
 \citet{Haluptzok2022}.}

\subsubsection*{InstructGPT}
\label{instructgpt}

Instruction fine-tuning and reinforcement learning from human feedback (RLHF) are now standard practices to improve LLM performance at following user instructions. 
\citet{Ouyang2022} introduced these techniques for instruction-following large scale models.

The first post-training enhancement is collecting human demonstrations of instruction following, and then fine-tuning the model on these demonstrations in a supervised fashion. The amount of compute used in fine-tuning is $\sim0.1\%$ of the amount used for pre-training.\footnote{The dataset consisted of $\sim 13000$ demonstrations, and the model was trained for 16 epochs. Assuming an average sequence length of 2048 tokens (GPT-3 context window size, see
\citet{Brown2020}) the number of tokens seen during fine-tuning is 
$\mathrm{13000\ast 16\ast 2048 \sim 426M}$. This is 0.14\% the number of tokens used for pre-training (300B, see \citet{Brown2020}). }

The second enhancement is collecting human comparisons between several generated responses and training a reward model on this data. This reward model learns to predict the rating that a human would give to a generated answer. Then, the reward model can be used to train the original model via reinforcement learning. This process takes $\sim0.2\%$ as much compute as pre-training.\footnote{The reward model has 6B parameters and is fine-tuned on 33k sequences for 1 epoch (see Appendix C.3 and Table 6 in \citet{Brown2020}). This gives $\mathrm{6\ast 6e9\ast 33e3\ast 2048 = 2.4e18}$ FLOP, five orders of magnitude less than the pre-training cost. Training via reinforcement learning takes 256k episodes. Assuming an average episode length of 2048, we get $\mathrm{256e3\ast 2048 = 5.24e8}$ tokens, which is 0.17\% of those used in pre-training. }

They compared the rate at which human judges preferred various models' responses to a supervised fine-tuning baseline. This ``win rate'' is shown in the following figure.

\begin{figure}[htb]\centering
 \includegraphics[width=0.45\textwidth]{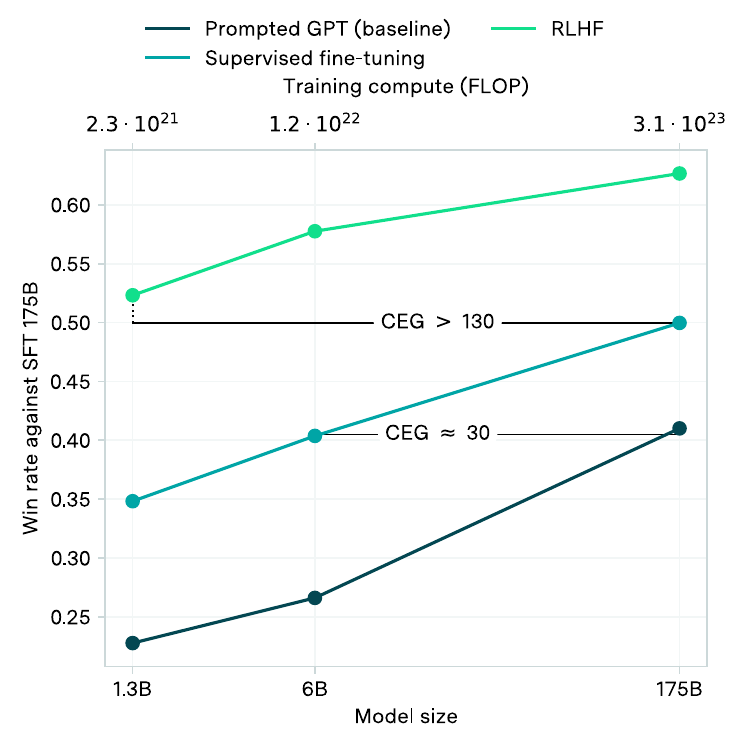}
 \caption*{Rater preference for GPT models fine-tuned on instructions and human feedback.\footnotemark}
\end{figure}
\footnotetext{Data from Figure 1 of \citet{Ouyang2022}}
 
Prompting GPT produces a larger improvement than scaling the model size by $130\times$. Fine-tuning further improves over this more than scaling by $30\times$, and RLHF further improves over fine-tuning more than scaling by $130\times$. Combining the benefits from supervised fine-tuning and RLHF (the two data enhancements), we estimate a total CEG of $>3900$.\footnote{Since all GPT-3 variants were trained on 300B tokens, the amount of training compute is proportional to the model size. $4\ast130=520$.}

This metric is overly optimistic since it is precisely the optimization objective of the reinforcement learning process. Evaluations on different NLP datasets show little or no improvement from GPT to InstructGPT. The NLP benchmarks with the largest improvements are TruthfulQA, CNN/DM and TLDR. In those three benchmarks, InstructGPT-1.3B performs better than GPT-3-175B, which corresponds to a CEG of $>130$.\footnote{Once again, the CEG is proportional to the increase in parameters, in this case $\mathrm{175B/1.3B = 130}$. }

\section{Limitations to quantifying performance gains via the CEG}
\label{sec4}

To quantify how much post-training enhancements increase model capabilities, we measured the compute-equivalent gain (CEG): the increase in pretraining compute that would be required to match the performance improvement from the post-training enhancement.

However, this metric can be difficult to both \textbf{measure} and \textbf{interpret}.

When measuring the CEG we would ideally compare models from the same family that were trained compute-optimally. But this is rarely the case in our examples.

With regards to interpretation, benchmarks may have been selected to exaggerate the effect of the post-training enhancements. In addition, a high CEG might not indicate that the post-training enhancement significantly improves performance, but instead indicate that additional training compute doesn't improve performance.

\subsection*{Difficulties in measuring the CEG}

In this paper, we did not run our own experiments to calculate the CEG. Instead, we used results from other papers. This made it difficult to obtain rigorous estimates, since our methodology can only provide bounds on the CEG or point estimates based on scaling laws fit from very few datapoints.

\subsubsection*{Variation with scale}

The performance improvement from a given enhancement might increase or decrease with the scale of the model (see \Cref{fig:scale_inv} for a toy example). For this reason, the CEG of an enhancement might not be the same at all scales. In this paper we only make use of the CEG to quantify the gain from each enhancement at a particular scale,\footnote{Although the scale is different for each enhancement. If all of them show increasing gains with scale, this would bias any comparison of CEGs between enhancements.} and avoid extrapolating these gains to larger-scale models.

This dependence of the CEG on scale introduces some potential ambiguity: we define the CEG as the additional compute that a non-enhanced model would need to match the performance of an enhanced one. But an equally valid alternative definition would be the reduction in compute that can be achieved with an enhanced model without reducing performance. If the CEG is not scale-invariant, these definitions will give different values.

Ultimately, both definitions capture the effect of a post-training enhancement on the performance scaling curve, but care should be taken  when comparing between them.

\subsubsection*{Models from different families}

In two cases we compare models from different families, introducing noise into the CEG estimates. To estimate the CEG, we compare the performance of two models: a smaller model with the enhancement and a larger model without it. However, sometimes the small and large models differ in other ways than the amount of pretraining compute (e.g. different architectures or different training data), and these confounding factors make the estimate of the CEG less reliable. This is the case for two out of our 13 estimates of the CEG.\footnote{For \hyperref[orca]{Orca} and \hyperref[toolformer]{Toolformer}.}

For example, we compared Orca (\citet{Mukherjee2023}), a fine-tuned 13B LLaMa model, with ChatGPT. These models have different architectures and training data, which adds uncertainty to our estimated CEG.

\subsubsection*{Suboptimally scaled models}

In many cases we compare models that are not trained compute-optimally, biasing the CEG estimate. Even when models come from the same family and differ only by the amount of pretraining compute, estimates of the CEG can still be misleading if the family of models were not trained compute-optimally. If the family of models \textit{had} been trained compute-optimally then the performance gain from increasing training compute would be different, and so the calculated CEG would be different.

For example, when estimating the CEG of chain-of-thought prompting, we compare the performance of PALM-540B to that of PALM-8B. But PALM-540B was undertrained relative to its size, and the performance gain would have been larger if both models had been trained compute-optimally. This means we overestimate the CEG.\footnote{Conversely, we would have underestimated the CEG if PALM-540B had been trained compute-optimally and PALM-8B had been over-trained.}
Indeed, in \Cref{apB} we got much lower estimates using data from the models in \citet{Lanham2023}.

\subsection*{Pitfalls in interpreting the CEG}
\subsubsection*{High CEG might indicate that more compute doesn't improve performance}

\begin{figure}[htb]\centering
 \includegraphics[width=0.45\textwidth]{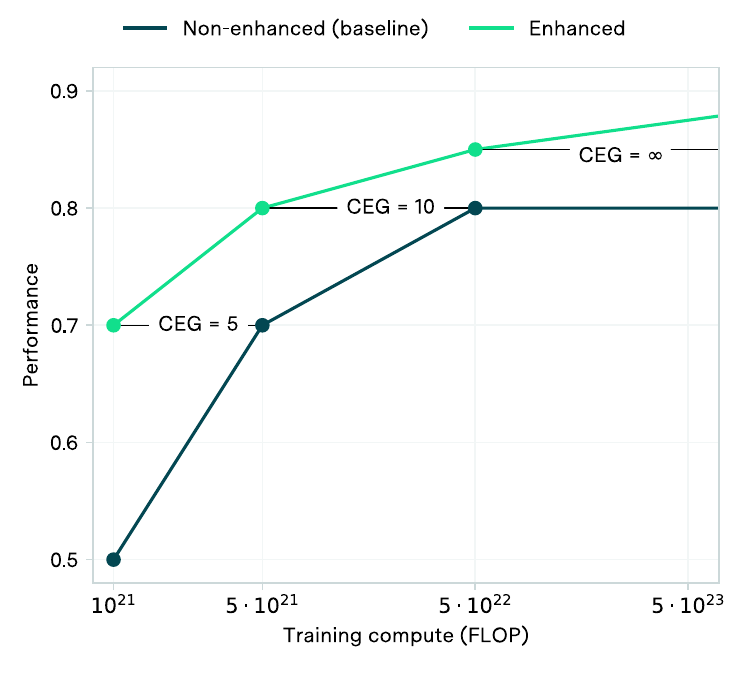}
\caption{Simulated example illustrating how the CEG of an enhancement can vary with scale. If scaling the baseline does not improve performance, the CEG stops being meaningful.}
 \label{fig:scale_inv}
\end{figure}

Even if we estimate the CEG using optimally scaled models from the same family, a high CEG on a dataset doesn't necessarily mean that the post-training enhancement is useful. It could also mean that for this particular dataset, the baseline of spending more compute on training is ineffective (see \Cref{fig:scale_inv}).

Relatedly, some post-training enhancements allow models to perform tasks that would be impossible for any model without it. In these cases, even if the enhancement greatly improves performance, the CEG is not meaningful. For example, many ``LLM agent'' scaffolds significantly increase the capabilities of language models by allowing them to execute code, store memories, or browse the web. Since no raw language model could perform these tasks, calculating CEG by comparing a smaller LLM agent to a larger raw language model is not appropriate. More promising would be to calculate the CEG by comparing two different agent architectures, but there is little available data for this.

\subsubsection*{Metric and dataset selection bias}

Authors may be tempted to report metrics and evaluate on datasets that exaggerate the size of their contributions. This will increase estimates of the CEG.

In addition, authors are generally more likely to compare their enhancement against a baseline of spending more compute -- thereby providing enough data to estimate a CEG -- if this emphasizes their method's strengths. These selection biases mean that the highest CEG numbers may be due to weak scaling baselines (see previous section), and may not transfer to other datasets.

\subsection*{Takeaway}

Due to these limitations, individual CEG estimates can be noisy. Nonetheless, we think that the estimates provide a convenient and intuitive measure of the effectiveness of post-training enhancements. Situations in which the CEG estimates are most misleading (such as inflated estimates in cases where compute scaling does not improve performance) can be identified experimentally, and the corresponding estimates can be corrected or excluded from the analysis. Finally, many uncertain estimates can still be informative in aggregate. In particular, they indicate that post-training enhancements can cheaply provide performance improvements that would require scaling pre-training by one or two orders of magnitude.

\section{Discussion}
\label{sec5}

This section comments on trends in the examples discussed above. It points out that multiple enhancements can be combined together, argues that post-training enhancements will continue to improve capabilities in the future, and suggests that enhancements could make models more dangerous.

\subsection*{Multiple post-training enhancements can be combined together}

We've seen many examples of multiple post-training enhancements being combined together:
\begin{itemize}
    \item \citet{Lewkowycz2022} combines data cleaning with majority voting in Minerva, reaching a CEG of 30 in STEM benchmarks and 2400 in math benchmarks.

    \item \citet{Nakano2021} combines tool-use with best-of-n in WebGPT, reaching a CEG of 220 in TruthfulQA.

    \item AI agents typically combine multiple post-training enhancements relating to prompting, scaffolding, and tool-use.

    \item \citet{Ouyang2022} combines supervised fine-tuning with RLHF in InstructGPT, reaching a CEG of 130 in some NLP benchmarks and 3900 in preference ratings.
\end{itemize}

The CEG achieved by these combined enhancements are the highest that we have observed, significantly above the median, which indicates large benefits from combining enhancements. However, we have reasons to doubt each of these individual estimates (see \Cref{sec3}) so we don't place much confidence in this conclusion.

When combining multiple post-training enhancements, it might be the case that the CEG of the combination is not the same as the product of the individual CEGs, as one might naively expect. We suspect there are diminishing returns to using more post-training enhancements for the same downstream task, such that the combined CEG becomes much lower than the product of individual CEGs. Unfortunately, we do not have enough data to study this question, as there are few studies that report results for both individual and combined enhancements. 

\subsection*{Post-training enhancements will probably continue to improve model capabilities into the future}

If more post-training enhancements continue being found, then a current generation model like GPT-4 or Llama 2 might, in the future, be significantly more capable in certain areas than it is today with only minor changes to the base model.

Some of the post-training enhancements discussed here seem like they might be ``one time gains''. For example, majority vote (where the model generates many candidate solutions and you submit the most common answer) and best-of-n (where multiple candidate solutions are evaluated and you submit the candidate with the highest evaluation score) are obvious techniques and there may not be significantly better ways to select candidate solutions.

Other post-training enhancements seem like they could be improved on significantly. Models could learn to use a wider variety of tools, and could get more practice in using them; we could discover new techniques for generating high-quality fine-tuning data (especially as AIs get better at generating synthetic data), and there may be significant further improvements to prompting techniques and agent architectures. Indeed, we've seen several examples of post-training enhancements that build on or displace previous enhancements:
\begin{itemize}
\item Chain of thought prompting has been extended and improved in multiple papers \cite{Wang2023, Press2022, Huang2022}.

\item Outcomes-based verification \cite{Cobbe2021}) was later improved upon with process-based verification \cite{Lightman2023}.

\item \citet{Li2022} used multiple fine-tuning and filtering techniques which iteratively improve the performance of AlphaCode.

\item The LATS agent \cite{Zhou2023} outperforms previous agents such as ReAct, Reflexion and Tree of Thoughts.
\end{itemize}

Larger models often benefit more from post-training enhancements than smaller models. For example, chain-of-thought has bigger benefits for larger models and small versions of Toolformer struggled to learn to use tools. This suggests that, as model size continues to increase, new post-training enhancements will become accessible. 

So we expect future post-training enhancements to contribute significantly to improvements in model capabilities.

\subsection*{Post-training enhancements have varied skill profiles}

Some post-training enhancements are fairly general. Chain-of-thought is an example (though it improves performance more in math and reasoning tasks), as are majority voting and best-of-n. Agent architectures enhance some skills (e.g. autonomy, planning and error-correction) that are useful across a wide range of domains.

Many post-training enhancements only improve performance in a narrow domain. For example, teaching the Toolformer \cite{Schick2023} to use a calculator only improves its performance on tasks involving arithmetic, and ``training a verifier to evaluate math solutions'' 
\cite{Cobbe2021, Lightman2023} only improves performance on math problem solving.

Even for these narrow enhancements, it often seems like similar post-training enhancements might lead to improvements in a different domain. For example, the capabilities enhanced by Toolformer depend on the tools that are used, and one could train verifiers on many tasks (e.g. on instructions for carrying out biological experiments). Indeed, \citet{Boiko2023} enhanced capabilities in a narrow domain by connecting an AI system with robotic hardware for synthesizing chemicals.

\section{Policy Implications}
\label{sec6}
 
In this section, we discuss implications of post-training enhancements relating to dangerous AI capabilities and compute governance.

Recent AI governance proposals \cite{Shevlane2023, Anderljung2023} recommend that frontier AI models be evaluated for dangerous capabilities before they are deployed. \citet{Shevlane2023} argue a model should be treated as highly dangerous ``if it has a capability profile that would be sufficient for extreme harm, assuming misuse and/or misalignment'', for example, if the model can gain access to weapons or acquire political influence, and appropriate defenses against these threats have not been implemented.
Models exhibiting highly dangerous capabilities then might not be deployed at all, or structured access might be provided \cite{Shevlane2023, Solaiman2023}, e.g. allowing reversible API access but not releasing the model weights. 

Post-training enhancements might enhance a model’s dangerous capabilities. This could happen as a consequence of enhancing capabilities in a general way, but also if a narrow enhancement specifically improves some dangerous capability. For example, the general agent from \citet{kinniment2023} researched someone on the Internet and then drafted a personalized phishing email (though it made several critical mistakes in the process). Meanwhile, the autonomous scientific research agent developed by \citet{Boiko2023} enhanced capabilities in chemical synthesis. Further enhancements in this domain could reach the point of being useful for manufacturing weapons.

This has a number of implications for dangerous capability evaluations. First, the \textit{best available post-training enhancements} should be used when evaluating a model’s dangerous capabilities, so that evaluations consider all the capabilities that could be unlocked by post-training enhancements rather than the capabilities of the pretrained model alone \cite{kinniment2023}.

Second, measurements of a model’s dangerous capabilities should be treated as a lower bound on what is possible, as future enhancements may increase those capabilities, perhaps significantly. Developers could err on the side of caution by incorporating a safety buffer into their deployment decisions (as in \citet{Anthropic2023}). Specifically, protective measures (like restricting deployment) would be triggered at lower capability levels than those that have been defined as concerning. This way, a model’s dangerous capabilities will remain unconcerning even with additional post-training enhancements. In practice, this could, for example, translate into lowering the success rate required to count a task as passed or the number of tasks a model must pass to consider it has a certain capability.\footnote{For example, \citet{Anthropic2023} counts a task as ``passed'' if the model succeeds 10\% of the time, and considers an evaluation threshold as met if at least 50\% of the tasks are passed. The organization presents these numbers as a conservative approach that accounts for a safety buffer.}

However, both the need for safety buffers and exhaustive use of enhancements during evaluations might be lower if developers monitor the model's dangerous capabilities along the value chain and respond to its evolution accordingly. More specifically, developers could supervise downstream uses by analyzing API inputs and outputs, extending existing practices \cite{OpenAI2023, Anthropic2023}. Then, developers could prevent or abort risky enhancements, for instance, through capability or feature restrictions and access frequency limits \cite{OBrien2023}.

Finally, post-training enhancements challenge proposals to govern frontier AI capabilities by regulating large training runs. Few actors can afford the large quantities of AI chips needed to train an frontier model from scratch\footnote{The amortized cost of compute for training GPT-4 is estimated to be between \$30 and \$90 million \cite{epoch2023aitrends}}. If systems for monitoring access to large quantities of AI chips were in place \cite{Egan2023}, a governance regime might identify training runs for frontier models and apply regulatory requirements to them \cite{Anderljung2023, Shavit2023}. However, this regime would not identify post-training enhancements with low compute costs, which are accessible to a wider group of actors. Thus governance focused on training should be extended to include mechanisms to govern post-training enhancements and deployment.

\section{Future Work}
\label{sec7}

Future research into post-training enhancements could improve estimates of the CEG, investigate the gains from combining multiple post-training enhancements together, and explore policy implications.

\subsection*{Improving estimates of the CEG}

We have only provided rough bounds and estimates of the CEG from these techniques. There are multiple ways to improve these estimates.

\subsubsection*{Collect more data points on each post-training enhancement}

We only used data from one paper for each post-training enhancement.

Researchers could collect data for a wider range of benchmarks for each enhancement. This could clarify the kind of capabilities that a given post-training enhancement can improve, how general these improvements are, and how consistent the CEG is across similar benchmarks (e.g. is the CEG much bigger on some math benchmarks than others?).

In addition, researchers could collect data for a wider range of model sizes. This could shed light on how the effect of an enhancement changes as models become larger. Researchers could also calculate scaling laws for benchmark performance with and without the enhancement and use these to calculate the CEG (see \Cref{apA1}).

\subsubsection*{Run experiments to calculate the CEG}

Even better, researchers could run controlled experiments: using the same model family for enhanced and non-enhanced models, training compute-optimally, and testing in pre-specified benchmarks. They could isolate the effect of the post-training enhancement and avoid many of the limitations of our methodology discussed in section 4.

\subsubsection*{Calculate the CEG for agent scaffolding}

As we mentioned in section 4, estimating the CEG from agent scaffolding is often difficult because usually bare LLMs simply can't perform the type of tasks that agents are designed to perform.

There are several possible approaches to mitigate this problem. Researchers could find more benchmarks in which bare LLMs can compete with agents, like HumanEval. But this would only work in a few domains. Researchers could also compare the performance of different agent architectures on the same agent benchmark. This would require innovations in benchmarking LLM agents, for example by building on 
\citet{Ruan2023}, 
\citet{Liu2023}, or 
\citet{Yang2023}.

\subsection*{Investigate the combined impact of multiple post-training enhancements on AI capabilities}

Some of the \textit{individual} post-training enhancements we mentioned, like chain-of-thought, have been intensely studied. However, the \textit{combined} effects of multiple enhancements is also relevant for policy. There are several ways to improve our understanding of these effects.

\subsubsection*{Study the diminishing returns to additional post-training enhancements}

It seems unlikely that the CEG of multiple enhancements is as high as the product of the individual CEGs, since this would quickly result in an extremely high CEG. So there are likely diminishing returns as the number of enhancements applied to a model increases. Researchers could study this by running controlled experiments with different combinations of enhancements.

Researchers could also study whether there is a ceiling to the total improvement you can get from post-training enhancements, no matter how many you apply. If policy makers could put a ceiling on the total improvement from future post-training enhancements, they could better predict how much more dangerous a model might become in the future.

\subsubsection*{Investigate the ``skill profile'' of post-training enhancements}

The gains from a single post-training enhancement are often narrow, but different enhancements might help with different tasks. This raises the question of how much more general the gains might be from multiple enhancements combined together. If policy makers knew the breadth of capability improvements that post-training enhancements provide in combination, they could again better predict how much more dangerous a model might become in the future.

\subsubsection*{Study the rate of improvement from post-training enhancements over time}

As more post-training enhancements are developed, and existing ones are improved, estimates of the CEG from the best enhancements will likely increase. Tracking how the CEG increases over time gives a rate of improvement for post-training enhancements. Comparing this rate of improvement with the rates of improvement from compute scaling or other algorithmic improvement, would give a better sense for how important post-training enhancements are overall for driving AI progress. It would also be informative for the design of safety buffers.

\subsection*{Policy questions}
The policy implications of post-training enhancements that we outlined in Section 6 raise some further questions.
\begin{itemize}
\item How can we quickly identify new and potentially dangerous post-training enhancements?

\item How can we prevent dangerous post-training enhancements from being applied to models?

\item Which actors have historically found improved post-training enhancements, and what resources were necessary to discover them?

\item How should we govern complex ecosystems composed of many interacting AI systems and software systems from multiple providers?
\end{itemize}

\section{Conclusion}
\label{sec8}

We have introduced a basic framework for quantifying the benefits and costs of post-training enhancements, most notably measuring the benefits via the CEG. We applied this framework to a representative collection of post-training enhancements and saw that while the performance improvements can be significant, the fine-tuning costs are typically very small compared to the cost of pre-training. It seems likely that new post-training enhancements will continue to improve AI capabilities, though it's unclear how much total room for further improvement there is. The potential for a wide range of actors to improve frontier AI capabilities poses a distinct challenge for AI governance.

\section*{Acknowledgments}

We would like to express our thanks to the people who have offered feedback and input on the ideas in this paper, including Tamay Besiroglu, Ege Erdil, Matthew Barnett, Jaime Sevilla, Girish Sastry, Sam Bowman, Toby Shevlane, Lewis Ho, Markus Anderljung, Sébastien Krier, Cullen O'Keefe, Holden Karnofsky, Roger Grosse, Hjalmar Wijk, Jacob Hilton, Jacob Steinhardt, Marius Hobbhahn, Lukas Finnveden, Josh You and Eli Lifland.

\bibliography{refs.bib}

\begin{thebibliography}{64}
\providecommand{\natexlab}[1]{#1}
\providecommand{\url}[1]{\texttt{#1}}
\expandafter\ifx\csname urlstyle\endcsname\relax
  \providecommand{\doi}[1]{doi: #1}\else
  \providecommand{\doi}{doi: \begingroup \urlstyle{rm}\Url}\fi

\bibitem[Sevilla et~al.(2022)Sevilla, Heim, Ho, Besiroglu, Hobbhahn, and Villalobos]{sevilla2022}
Jaime Sevilla, Lennart Heim, Anson Ho, Tamay Besiroglu, Marius Hobbhahn, and Pablo Villalobos.
\newblock Compute trends across three eras of machine learning, 2022.
\newblock URL \url{https://arxiv.org/abs/2202.05924}.

\bibitem[Villalobos et~al.(2022)Villalobos, Sevilla, Heim, Besiroglu, Hobbhahn, and Ho]{Villalobos2022}
Pablo Villalobos, Jaime Sevilla, Lennart Heim, Tamay Besiroglu, Marius Hobbhahn, and Anson Ho.
\newblock Will we run out of data? an analysis of the limits of scaling datasets in machine learning, 2022.
\newblock URL \url{https://arxiv.org/abs/2211.04325}.

\bibitem[Hernandez and Brown(2020)]{Hernandez2020}
Danny Hernandez and Tom~B Brown.
\newblock Measuring the algorithmic efficiency of neural networks, 2020.
\newblock URL \url{https://arxiv.org/abs/2005.04305}.

\bibitem[Erdil and Besiroglu(2023)]{erdil2023algorithmic}
Ege Erdil and Tamay Besiroglu.
\newblock Algorithmic progress in computer vision, 2023.
\newblock URL \url{https://arxiv.org/abs/2212.05153}.

\bibitem[Srivastava et~al.(2023)Srivastava, Rastogi, Rao, Shoeb, Abid, and {Adam Fisch, et al. (BIG-bench collaboration)}]{Srivastava2023}
Aarohi Srivastava, Abhinav Rastogi, Abhishek Rao, Abu Awal~Md Shoeb, Abubakar Abid, and {Adam Fisch, et al. (BIG-bench collaboration)}.
\newblock Beyond the imitation game: Quantifying and extrapolating the capabilities of language models.
\newblock \emph{Transactions on Machine Learning Research}, 2023.
\newblock ISSN 2835-8856.
\newblock URL \url{https://openreview.net/forum?id=uyTL5Bvosj}.

\bibitem[Villalobos and Atkinson(2023)]{Villalobos2023}
Pablo Villalobos and David Atkinson.
\newblock Trading off compute in training and inference, 2023.
\newblock URL \url{https://epochai.org/blog/trading-off-compute-in-training-and-inference}.
\newblock Accessed: 2023-11-10.

\bibitem[Anderljung et~al.(2023)Anderljung, Barnhart, Korinek, Leung, OKeefe, Whittlestone, Avin, Brundage, CassBeggs, Chang, Collins, Fist, Hayes, Ho, Hooker, Horvitz, Kolt, Shavit, Siddarth, Trager, and Wolf]{Anderljung2023}
Markus Anderljung, Joslyn Barnhart, Anton Korinek, Jade Leung, Cullen OKeefe, Jess Whittlestone, Shahar Avin, Miles Brundage, Justin Bullock~Duncan CassBeggs, Ben Chang, Tantum Collins, Tim Fist, Gillian Hadfield~Alan Hayes, Lewis Ho, Sara Hooker, Eric Horvitz, Noam Kolt, Jonas Schuett~Yonadav Shavit, Divya Siddarth, Robert Trager, and Kevin Wolf.
\newblock Frontier ai regulation: Managing emerging risks to public safety, 2023.
\newblock URL \url{https://arxiv.org/abs/2307.03718}.

\bibitem[Mialon et~al.(2023)Mialon, Dessi, Lomeli, Nalmpantis, Pasunuru, Raileanu, Roziere, Schick, DwivediYu, Celikyilmaz, Grave, LeCun, and Scialom]{Mialon2023}
Gregoire Mialon, Roberto Dessi, Maria Lomeli, Christoforos Nalmpantis, Ram Pasunuru, Roberta Raileanu, Baptiste Roziere, Timo Schick, Jane DwivediYu, Asli Celikyilmaz, Edouard Grave, Yann LeCun, and Thomas Scialom.
\newblock Augmented language models: a survey, 2023.
\newblock URL \url{https://arxiv.org/abs/2302.07842}.

\bibitem[Hendrycks et~al.(2021)Hendrycks, Burns, Kadavath, Arora, Basart, Tang, Song, and Steinhardt]{Hendrycks2021}
Dan Hendrycks, Collin Burns, Saurav Kadavath, Akul Arora, Steven Basart, Eric Tang, Dawn Song, and Jacob Steinhardt.
\newblock Measuring mathematical problem solving with the math dataset, 2021.
\newblock URL \url{https://arxiv.org/abs/2103.03874}.

\bibitem[Brown et~al.(2020)Brown, Mann, Ryder, Subbiah, Kaplan, Dhariwal, Neelakantan, Shyam, Askell, Agarwal, HerbertVoss, Krueger, Henighan, Child, Ramesh, Ziegler, Wu, Winter, Hesse, Chen, Sigler, Litwin, Gray, Chess, Clark, Berner, McCandlish, Radford, Sutskever, and Amodei]{Brown2020}
Tom~B Brown, Benjamin Mann, Nick Ryder, Melanie Subbiah, Jared Kaplan, Prafulla Dhariwal, Arvind Neelakantan, Pranav Shyam, Girish Sastry~Amanda Askell, Sandhini Agarwal, Ariel HerbertVoss, Gretchen Krueger, Tom Henighan, Rewon Child, Aditya Ramesh, Daniel~M Ziegler, Jeffrey Wu, Clemens Winter, Christopher Hesse, Mark Chen, Eric Sigler, Mateusz Litwin, Scott Gray, Benjamin Chess, Jack Clark, Christopher Berner, Sam McCandlish, Alec Radford, Ilya Sutskever, and Dario Amodei.
\newblock Language models are few-shot learners, 2020.
\newblock URL \url{https://arxiv.org/abs/2005.14165}.

\bibitem[Press et~al.(2022)Press, Zhang, Min, Schmidt, Smith, and Lewis]{Press2022}
Ofir Press, Muru Zhang, Sewon Min, Ludwig Schmidt, Noah~A Smith, and Mike Lewis.
\newblock Measuring and narrowing the compositionality gap in language models, 2022.
\newblock URL \url{https://arxiv.org/abs/2210.03350}.

\bibitem[Wei et~al.(2022)Wei, Wang, Schuurmans, Bosma, Xia, Chi, Le, and Zhou]{Wei2022}
Jason Wei, Xuezhi Wang, Dale Schuurmans, Maarten Bosma, Brian Ichter~Fei Xia, Ed~Chi, Quoc Le, and Denny Zhou.
\newblock Chain-of-thought prompting elicits reasoning in large language models, 2022.
\newblock URL \url{https://arxiv.org/abs/2201.11903}.

\bibitem[Lewkowycz et~al.(2022)Lewkowycz, Andreassen, Dohan, Dyer, Michalewski, Ramasesh, Slone, Anil, Schlag, GutmanSolo, Wu, Neyshabur, GurAri, and Misra]{Lewkowycz2022}
Aitor Lewkowycz, Anders Andreassen, David Dohan, Ethan Dyer, Henryk Michalewski, Vinay Ramasesh, Ambrose Slone, Cem Anil, Imanol Schlag, Theo GutmanSolo, Yuhuai Wu, Behnam Neyshabur, Guy GurAri, and Vedant Misra.
\newblock Solving quantitative reasoning problems with language models, 2022.
\newblock URL \url{https://arxiv.org/abs/2206.14858}.

\bibitem[Parisi et~al.(2022)Parisi, Zhao, and Fiedel]{Parisi2022}
Aaron Parisi, Yao Zhao, and Noah Fiedel.
\newblock Talm: Tool augmented language models, 2022.
\newblock URL \url{https://arxiv.org/abs/2205.12255}.

\bibitem[Li et~al.(2022)Li, Choi, Chung, Kushman, Schrittwieser, Leblond, Eccles, Keeling, Lago, Hubert, Choy, de~Masson~dAutume, Babuschkin, Chen, Huang, Welbl, Gowal, Cherepanov, Molloy, Mankowitz, Kohli, de~Freitas, Kavukcuoglu, and Vinyals]{Li2022}
Yujia Li, David Choi, Junyoung Chung, Nate Kushman, Julian Schrittwieser, Remi Leblond, Tom Eccles, James Keeling, Felix Gimeno Agustin~Dal Lago, Thomas Hubert, Peter Choy, Cyprien de~Masson~dAutume, Igor Babuschkin, Xinyun Chen, PoSen Huang, Johannes Welbl, Sven Gowal, Alexey Cherepanov, James Molloy, Daniel~J Mankowitz, Esme Sutherland Robson~Pushmeet Kohli, Nando de~Freitas, Koray Kavukcuoglu, and Oriol Vinyals.
\newblock Competition-level code generation with alphacode, 2022.
\newblock URL \url{https://arxiv.org/abs/2203.07814}.

\bibitem[Wang et~al.(2023{\natexlab{a}})Wang, Xu, Lan, Hu, Lan, Lee, and Lim]{Wang2023}
Lei Wang, Wanyu Xu, Yihuai Lan, Zhiqiang Hu, Yunshi Lan, Roy~KaWei Lee, and EePeng Lim.
\newblock Plan-and-solve prompting: Improving zero-shot chain-of-thought reasoning by large language models, 2023{\natexlab{a}}.
\newblock URL \url{https://arxiv.org/abs/2305.04091}.

\bibitem[Huang et~al.(2022)Huang, Gu, Hou, Wu, Yu, and Han]{Huang2022}
Jiaxin Huang, Shixiang~Shane Gu, Le~Hou, Yuexin Wu, Xuezhi Wang~Hongkun Yu, and Jiawei Han.
\newblock Large language models can self-improve, 2022.
\newblock URL \url{https://arxiv.org/abs/2210.11610}.

\bibitem[Schick et~al.(2023)Schick, DwivediYu, Dessi, Raileanu, Lomeli, Zettlemoyer, Cancedda, and Scialom]{Schick2023}
Timo Schick, Jane DwivediYu, Roberto Dessi, Roberta Raileanu, Maria Lomeli, Luke Zettlemoyer, Nicola Cancedda, and Thomas Scialom.
\newblock Toolformer: Language models can teach themselves to use tools, 2023.
\newblock URL \url{https://arxiv.org/abs/2302.04761}.

\bibitem[Kinniment et~al.(2023)Kinniment, Koba~Sato, Du, Goodrich, Hasin, Chan, Miles, Lin, Wijk, Burget, Ho, Barnes, and Christiano]{kinniment2023}
Megan Kinniment, Lucas~Jun Koba~Sato, Haoxing Du, Brian Goodrich, Max Hasin, Lawrence Chan, Luke~Harold Miles, Tao~R Lin, Hjalmar Wijk, Joel Burget, Aaron Ho, Elizabeth Barnes, and Paul Christiano.
\newblock Evaluating language-model agents on realistic autonomous tasks, July 2023.
\newblock URL \url{https://evals.alignment.org/language-model-pilot-report}.

\bibitem[Boiko et~al.(2023)Boiko, MacKnight, and Gomes]{Boiko2023}
Daniil~A Boiko, Robert MacKnight, and Gabe Gomes.
\newblock Emergent autonomous scientific research capabilities of large language models, 2023.
\newblock URL \url{https://arxiv.org/abs/2304.05332}.

\bibitem[Epoch(2023)]{epoch2023aitrends}
Epoch.
\newblock Key trends and figures in machine learning, 2023.
\newblock URL \url{https://epochai.org/trends}.
\newblock Accessed: 2023-11-30.

\bibitem[Hoffmann et~al.(2022)Hoffmann, Borgeaud, Mensch, Cai, Rutherford, de~Las~Casas, Welbl, Clark, Hennigan, Noland, van~den Driessche, Damoc, Guy, Osindero, Simonyan, Elsen, Rae, Vinyals, and Sifre]{Hoffmann2022}
Jordan Hoffmann, Sebastian Borgeaud, Arthur Mensch, Elena Buchatskaya~Trevor Cai, Eliza Rutherford, Diego de~Las~Casas, Lisa Anne Hendricks~Johannes Welbl, Aidan Clark, Tom Hennigan, Eric Noland, Katie Millican~George van~den Driessche, Bogdan Damoc, Aurelia Guy, Simon Osindero, Karen Simonyan, Erich Elsen, Jack~W Rae, Oriol Vinyals, and Laurent Sifre.
\newblock Training compute-optimal large language models, 2022.
\newblock URL \url{https://arxiv.org/abs/2203.15556}.

\bibitem[Henighan et~al.(2020)Henighan, Kaplan, Katz, Chen, Hesse, Jackson, Jun, Brown, Dhariwal, Gray, Hallacy, Mann, Radford, Ramesh, Ryder, Ziegler, Schulman, Amodei, and McCandlish]{Henighan2020}
Tom Henighan, Jared Kaplan, Mor Katz, Mark Chen, Christopher Hesse, Jacob Jackson, Heewoo Jun, Tom~B. Brown, Prafulla Dhariwal, Scott Gray, Chris Hallacy, Benjamin Mann, Alec Radford, Aditya Ramesh, Nick Ryder, Daniel~M. Ziegler, John Schulman, Dario Amodei, and Sam McCandlish.
\newblock Scaling laws for autoregressive generative modeling, 2020.
\newblock URL \url{https://arxiv.org/abs/2010.14701}.

\bibitem[Hilton et~al.(2023)Hilton, Tang, and Schulman]{Hilton2023}
Jacob Hilton, Jie Tang, and John Schulman.
\newblock Scaling laws for single-agent reinforcement learning, 2023.
\newblock URL \url{https://arxiv.org/abs/2301.13442}.

\bibitem[Cottier(2023)]{cottier2023}
Ben Cottier.
\newblock Trends in the dollar training cost of machine learning systems, 2023.
\newblock URL \url{https://epochai.org/blog/trends-in-the-dollar-training-cost-of-machine-learning-systems}.
\newblock Accessed: 2023-11-10.

\bibitem[Wang and Komatsuzaki(2021)]{Komatsuzaki2021}
Ben Wang and Aran Komatsuzaki.
\newblock {GPT-J-6B: A 6 Billion Parameter Autoregressive Language Model}, May 2021.
\newblock URL \url{https://github.com/kingoflolz/mesh-transformer-jax}.

\bibitem[Zhang et~al.(2022)Zhang, Roller, Goyal, Artetxe, Chen, Dewan, Diab, Li, Lin, Mihaylov, Ott, Shleifer, Shuster, Simig, Koura, Sridhar, Wang, and Zettlemoyer]{Zhang2022}
Susan Zhang, Stephen Roller, Naman Goyal, Mikel Artetxe, Moya Chen~Shuohui Chen, Christopher Dewan, Mona Diab, Xian Li, Xi~Victoria Lin, Todor Mihaylov, Myle Ott, Sam Shleifer, Kurt Shuster, Daniel Simig, Punit~Singh Koura, Anjali Sridhar, Tianlu Wang, and Luke Zettlemoyer.
\newblock Opt: Open pre-trained transformer language models, 2022.
\newblock URL \url{https://arxiv.org/abs/2205.01068}.

\bibitem[Nakano et~al.(2021)Nakano, Hilton, Balaji, Wu, Kim, Hesse, Jain, Kosaraju, Saunders, Jiang, Cobbe, Eloundou, Krueger, Button, Knight, Chess, and Schulman]{Nakano2021}
Reiichiro Nakano, Jacob Hilton, Suchir Balaji, Jeff Wu, Long Ouyang~Christina Kim, Christopher Hesse, Shantanu Jain, Vineet Kosaraju, William Saunders, Xu~Jiang, Karl Cobbe, Tyna Eloundou, Gretchen Krueger, Kevin Button, Matthew Knight, Benjamin Chess, and John Schulman.
\newblock Webgpt: Browser-assisted question-answering with human feedback, 2021.
\newblock URL \url{https://arxiv.org/abs/2112.09332}.

\bibitem[Borgeaud et~al.(2021)Borgeaud, Mensch, Hoffmann, Cai, Rutherford, Millican, van~den Driessche, Damoc, Clark, de~Las~Casas, Guy, Ring, Hennigan, Huang, Maggiore, Jones, Cassirer, Brock, Paganini, Irving, Vinyals, Osindero, Simonyan, Rae, Elsen, and Sifre]{Borgeaud2021}
Sebastian Borgeaud, Arthur Mensch, Jordan Hoffmann, Trevor Cai, Eliza Rutherford, Katie Millican, George van~den Driessche, JeanBaptiste Lespiau~Bogdan Damoc, Aidan Clark, Diego de~Las~Casas, Aurelia Guy, Jacob Menick~Roman Ring, Tom Hennigan, Saffron Huang, Loren Maggiore, Chris Jones, Albin Cassirer, Andy Brock, Michela Paganini, Geoffrey Irving, Oriol Vinyals, Simon Osindero, Karen Simonyan, Jack~W Rae, Erich Elsen, and Laurent Sifre.
\newblock Improving language models by retrieving from trillions of tokens, 2021.
\newblock URL \url{https://arxiv.org/abs/2112.04426}.

\bibitem[Chowdhery et~al.(2022)Chowdhery, Narang, Devlin, Bosma, Mishra, Roberts, Barham, Chung, Sutton, Gehrmann, Schuh, Shi, Tsvyashchenko, Maynez, Rao, Barnes, Tay, Shazeer, Prabhakaran, Reif, Du, Hutchinson, Pope, Bradbury, Austin, Isard, Gur-Ari, Yin, Duke, Levskaya, Ghemawat, Dev, Michalewski, Garcia, Misra, Robinson, Fedus, Zhou, Ippolito, Luan, Lim, Zoph, Spiridonov, Sepassi, Dohan, Agrawal, Omernick, Dai, Pillai, Pellat, Lewkowycz, Moreira, Child, Polozov, Lee, Zhou, Wang, Saeta, Diaz, Firat, Catasta, Wei, Meier-Hellstern, Eck, Dean, Petrov, and Fiedel]{Chowdhery2022}
Aakanksha Chowdhery, Sharan Narang, Jacob Devlin, Maarten Bosma, Gaurav Mishra, Adam Roberts, Paul Barham, Hyung~Won Chung, Charles Sutton, Sebastian Gehrmann, Parker Schuh, Kensen Shi, Sasha Tsvyashchenko, Joshua Maynez, Abhishek Rao, Parker Barnes, Yi~Tay, Noam Shazeer, Vinodkumar Prabhakaran, Emily Reif, Nan Du, Ben Hutchinson, Reiner Pope, James Bradbury, Jacob Austin, Michael Isard, Guy Gur-Ari, Pengcheng Yin, Toju Duke, Anselm Levskaya, Sanjay Ghemawat, Sunipa Dev, Henryk Michalewski, Xavier Garcia, Vedant Misra, Kevin Robinson, Liam Fedus, Denny Zhou, Daphne Ippolito, David Luan, Hyeontaek Lim, Barret Zoph, Alexander Spiridonov, Ryan Sepassi, David Dohan, Shivani Agrawal, Mark Omernick, Andrew~M. Dai, Thanumalayan~Sankaranarayana Pillai, Marie Pellat, Aitor Lewkowycz, Erica Moreira, Rewon Child, Oleksandr Polozov, Katherine Lee, Zongwei Zhou, Xuezhi Wang, Brennan Saeta, Mark Diaz, Orhan Firat, Michele Catasta, Jason Wei, Kathy Meier-Hellstern, Douglas Eck, Jeff Dean, Slav Petrov, and Noah Fiedel.
\newblock Palm: Scaling language modeling with pathways, 2022.

\bibitem[Lanham et~al.(2023)Lanham, Chen, Radhakrishnan, Steiner, Denison, Hernandez, Li, Durmus, Hubinger, Kernion, Lukovsiute, Nguyen, Cheng, Joseph, Schiefer, Rausch, Larson, Kundu, Kadavath, Yang, Henighan, Maxwell, TelleenLawton, Hume, HatfieldDodds, Kaplan, Brauner, Bowman, and Perez]{Lanham2023}
Tamera Lanham, Anna Chen, Ansh Radhakrishnan, Benoit Steiner, Carson Denison, Danny Hernandez, Dustin Li, Esin Durmus, Evan Hubinger, Jackson Kernion, Kamile Lukovsiute, Karina Nguyen, Newton Cheng, Nicholas Joseph, Nicholas Schiefer, Oliver Rausch, Robin Larson, Sam McCandlish~Sandipan Kundu, Saurav Kadavath, Shannon Yang, Thomas Henighan, Timothy Maxwell, Timothy TelleenLawton, Tristan Hume, Zac HatfieldDodds, Jared Kaplan, Jan Brauner, Samuel~R Bowman, and Ethan Perez.
\newblock Measuring faithfulness in chain-of-thought reasoning, 2023.
\newblock URL \url{https://arxiv.org/abs/2307.13702}.

\bibitem[Nye et~al.(2021)Nye, Andreassen, GurAri, Austin, Bieber, Dohan, Lewkowycz, Luan, Sutton, and Odena]{Nye2021}
Maxwell Nye, Anders~Johan Andreassen, Guy GurAri, Henryk Michalewski~Jacob Austin, David Bieber, David Dohan, Aitor Lewkowycz, Maarten Bosma~David Luan, Charles Sutton, and Augustus Odena.
\newblock Show your work: Scratchpads for intermediate computation with language models, 2021.
\newblock URL \url{https://arxiv.org/abs/2112.00114}.

\bibitem[Yao et~al.(2023)Yao, Yu, Zhao, Shafran, Cao, and Narasimhan]{Yao2023}
Shunyu Yao, Dian Yu, Jeffrey Zhao, Izhak Shafran, Thomas L Griffiths~Yuan Cao, and Karthik Narasimhan.
\newblock Tree of thoughts: Deliberate problem solving with large language models, 2023.
\newblock URL \url{https://arxiv.org/abs/2305.10601}.

\bibitem[Zelikman et~al.(2023)Zelikman, Huang, Poesia, Goodman, and Haber]{Zelikman2023}
Eric Zelikman, Qian Huang, Gabriel Poesia, Noah~D. Goodman, and Nick Haber.
\newblock Parsel: Algorithmic reasoning with language models by composing decompositions, 2023.
\newblock URL \url{https://arxiv.org/abs/2212.10561}.

\bibitem[{Significant Gravitas}(2023)]{Significant_Gravitas_AutoGPT}
{Significant Gravitas}.
\newblock {AutoGPT}, 2023.
\newblock URL \url{https://github.com/Significant-Gravitas/AutoGPT}.

\bibitem[Mowshowitz(2023)]{Mowshowitz2023}
Zvi Mowshowitz.
\newblock On autogpt, 2023.
\newblock URL \url{https://www.lesswrong.com/posts/566kBoPi76t8KAkoD/on-autogpt}.
\newblock [Accessed 10-11-2023].

\bibitem[Wang et~al.(2023{\natexlab{b}})Wang, Ma, Feng, Zhang, Yang, Zhang, Chen, Tang, Chen, Lin, Zhao, Wei, and Wen]{Wang2023c}
Lei Wang, Chen Ma, Xueyang Feng, Zeyu Zhang, Hao Yang, Jingsen Zhang, Zhiyuan Chen, Jiakai Tang, Xu~Chen, Yankai Lin, Wayne~Xin Zhao, Zhewei Wei, and Ji-Rong Wen.
\newblock A survey on large language model based autonomous agents, 2023{\natexlab{b}}.
\newblock URL \url{https://arxiv.org/abs/2308.11432}.

\bibitem[Sumers et~al.(2023)Sumers, Yao, Narasimhan, and Griffiths]{Sumers2023}
Theodore~R Sumers, Shunyu Yao, Karthik Narasimhan, and Thomas~L Griffiths.
\newblock Cognitive architectures for language agents, 2023.
\newblock URL \url{https://arxiv.org/abs/2309.02427}.

\bibitem[Weng(2023)]{Weng2023}
Lilian Weng.
\newblock Llm-powered autonomous agents.
\newblock \emph{lilianweng.github.io}, 2023.
\newblock URL \url{https://lilianweng.github.io/posts/2023-06-23-agent/}.

\bibitem[Shinn et~al.(2023)Shinn, Cassano, Labash, Gopinath, Narasimhan, and Yao]{Shinn2023}
Noah Shinn, Federico Cassano, Beck Labash, Ashwin Gopinath, Karthik Narasimhan, and Shunyu Yao.
\newblock Reflexion: Language agents with verbal reinforcement learning, 2023.
\newblock URL \url{https://arxiv.org/abs/2303.11366}.

\bibitem[Wang et~al.(2023{\natexlab{c}})Wang, Xie, Jiang, Mandlekar, Xiao, Zhu, Fan, and Anandkumar]{Wang2023b}
Guanzhi Wang, Yuqi Xie, Yunfan Jiang, Ajay Mandlekar, Chaowei Xiao, Yuke Zhu, Linxi Fan, and Anima Anandkumar.
\newblock Voyager: An open-ended embodied agent with large language models, 2023{\natexlab{c}}.
\newblock URL \url{https://arxiv.org/abs/2305.16291}.

\bibitem[Park et~al.(2023)Park, OBrien, Cai, Morris, Liang, and Bernstein]{Park2023}
Joon~Sung Park, Joseph~C OBrien, Carrie~J Cai, Meredith~Ringel Morris, Percy Liang, and Michael~S Bernstein.
\newblock Generative agents: Interactive simulacra of human behavior, 2023.
\newblock URL \url{https://arxiv.org/abs/2304.03442}.

\bibitem[Zhou et~al.(2023)Zhou, Yan, Shlapentokh-Rothman, Wang, and Wang]{Zhou2023}
Andy Zhou, Kai Yan, Michal Shlapentokh-Rothman, Haohan Wang, and Yu-Xiong Wang.
\newblock Language agent tree search unifies reasoning acting and planning in language models, 2023.
\newblock URL \url{https://arxiv.org/abs/2310.04406}.

\bibitem[Epoch(2022)]{epochMachineLearningData2022}
Epoch.
\newblock Parameter, compute and data trends in machine learning.
\newblock https://epochai.org/data/pcd, 2022.

\bibitem[Chen et~al.(2023)Chen, Shu, Shareghi, Collier, Narasimhan, and Yao]{Chen2023}
Baian Chen, Chang Shu, Ehsan Shareghi, Nigel Collier, Karthik Narasimhan, and Shunyu Yao.
\newblock Fireact: Toward language agent fine-tuning, 2023.
\newblock URL \url{https://arxiv.org/abs/2310.05915}.

\bibitem[Cobbe et~al.(2021)Cobbe, Kosaraju, Bavarian, Chen, Kaiser, Plappert, Tworek, Hilton, Nakano, Hesse, and Schulman]{Cobbe2021}
Karl Cobbe, Vineet Kosaraju, Mohammad Bavarian, Mark Chen, Heewoo Jun~Lukasz Kaiser, Matthias Plappert, Jerry Tworek, Jacob Hilton, Reiichiro Nakano, Christopher Hesse, and John Schulman.
\newblock Training verifiers to solve math word problems, 2021.
\newblock URL \url{https://arxiv.org/abs/2110.14168}.

\bibitem[Lightman et~al.(2023)Lightman, Kosaraju, Burda, Edwards, Baker, Lee, Leike, Schulman, Sutskever, and Cobbe]{Lightman2023}
Hunter Lightman, Vineet Kosaraju, Yura Burda, Harri Edwards, Bowen Baker, Teddy Lee, Jan Leike, John Schulman, Ilya Sutskever, and Karl Cobbe.
\newblock Let's verify step by step, 2023.
\newblock URL \url{https://arxiv.org/abs/2305.20050}.

\bibitem[Mukherjee et~al.(2023)Mukherjee, Mitra, Jawahar, Palangi, and Awadallah]{Mukherjee2023}
Subhabrata Mukherjee, Arindam Mitra, Ganesh Jawahar, Sahaj Agarwal~Hamid Palangi, and Ahmed Awadallah.
\newblock Orca: Progressive learning from complex explanation traces of gpt-4, 2023.
\newblock URL \url{https://arxiv.org/abs/2306.02707}.

\bibitem[Gudibande et~al.(2023)Gudibande, Wallace, Snell, Geng, Abbeel, Levine, and Song]{Gudibande2023}
Arnav Gudibande, Eric Wallace, Charlie Snell, Xinyang Geng, Hao Liu~Pieter Abbeel, Sergey Levine, and Dawn Song.
\newblock The false promise of imitating proprietary llms, 2023.
\newblock URL \url{https://arxiv.org/abs/2305.15717}.

\bibitem[Haluptzok et~al.(2022)Haluptzok, Bowers, and Kalai]{Haluptzok2022}
Patrick Haluptzok, Matthew Bowers, and Adam~Tauman Kalai.
\newblock Language models can teach themselves to program better, 2022.
\newblock URL \url{https://arxiv.org/abs/2207.14502}.

\bibitem[Gao et~al.(2020)Gao, Biderman, Black, Golding, Foster, Phang, He, Thite, Nabeshima, Presser, and Leahy]{Gao2020}
Leo Gao, Stella Biderman, Sid Black, Laurence Golding, Travis Hoppe~Charles Foster, Jason Phang, Horace He, Anish Thite, Noa Nabeshima, Shawn Presser, and Connor Leahy.
\newblock The pile: An 800gb dataset of diverse text for language modeling, 2020.
\newblock URL \url{https://arxiv.org/abs/2101.00027}.

\bibitem[Black et~al.(2021)Black, Leo, Wang, Leahy, and Biderman]{Black2021}
Sid Black, Gao Leo, Phil Wang, Connor Leahy, and Stella Biderman.
\newblock {GPT-Neo: Large Scale Autoregressive Language Modeling with Mesh-Tensorflow}, August 2021.
\newblock URL \url{https://doi.org/10.5281/zenodo.5297715}.

\bibitem[Ouyang et~al.(2022)Ouyang, Wu, Jiang, Almeida, Mishkin, Zhang, Agarwal, Slama, Ray, Schulman, Hilton, Kelton, Miller, Simens, Askell, Welinder, Christiano, Leike, and Lowe]{Ouyang2022}
Long Ouyang, Jeff Wu, Xu~Jiang, Diogo Almeida, Carroll L Wainwright~Pamela Mishkin, Chong Zhang, Sandhini Agarwal, Katarina Slama, Alex Ray, John Schulman, Jacob Hilton, Fraser Kelton, Luke Miller, Maddie Simens, Amanda Askell, Peter Welinder, Paul Christiano, Jan Leike, and Ryan Lowe.
\newblock Training language models to follow instructions with human feedback, 2022.
\newblock URL \url{https://arxiv.org/abs/2203.02155}.

\bibitem[Shevlane et~al.(2023)Shevlane, Farquhar, Garfinkel, Phuong, Whittlestone, Leung, Kokotajlo, Marchal, Anderljung, Kolt, Ho, Siddarth, Avin, Kim, Gabriel, Bolina, Clark, Bengio, Christiano, and Dafoe]{Shevlane2023}
Toby Shevlane, Sebastian Farquhar, Ben Garfinkel, Mary Phuong, Jess Whittlestone, Jade Leung, Daniel Kokotajlo, Nahema Marchal, Markus Anderljung, Noam Kolt, Lewis Ho, Divya Siddarth, Shahar Avin, Will Hawkins~Been Kim, Iason Gabriel, Vijay Bolina, Jack Clark, Yoshua Bengio, Paul Christiano, and Allan Dafoe.
\newblock Model evaluation for extreme risks, 2023.
\newblock URL \url{https://arxiv.org/abs/2305.15324}.

\bibitem[Solaiman(2023)]{Solaiman2023}
Irene Solaiman.
\newblock The gradient of generative ai release: Methods and considerations, 2023.
\newblock URL \url{https://arxiv.org/abs/2302.04844}.

\bibitem[Anthropic(2023)]{Anthropic2023}
Anthropic.
\newblock Anthropic's responsible scaling policy, 2023.
\newblock URL \url{https://www-files.anthropic.com/production/files/responsible-scaling-policy-1.0.pdf}.
\newblock [Accessed 20-11-2023].

\bibitem[OpenAI(2023)]{OpenAI2023}
OpenAI.
\newblock Gpt-4 system card, 2023.
\newblock URL \url{https://cdn.openai.com/papers/gpt-4-system-card.pdf}.
\newblock [Accessed 20-11-2023].

\bibitem[O'Brien et~al.(2023)O'Brien, Ee, and Williams]{OBrien2023}
Joe O'Brien, Shaun Ee, and Zoe Williams.
\newblock Deployment corrections: An incident response framework for frontier ai models, 2023.
\newblock URL \url{https://static1.squarespace.com/static/64edf8e7f2b10d716b5ba0e1/t/651c397fc04af033499df9f8/1696348544356/Deployment+corrections_+an+incident+response+framework+for+frontier+AI+models.pdf}.
\newblock [Accessed 20-11-2023].

\bibitem[Egan and Heim(2023)]{Egan2023}
Janet Egan and Lennart Heim.
\newblock Oversight for frontier ai through a know-your-customer scheme for compute providers, 2023.
\newblock URL \url{https://arxiv.org/abs/2310.13625}.

\bibitem[Shavit(2023)]{Shavit2023}
Yonadav Shavit.
\newblock What does it take to catch a chinchilla? verifying rules on large-scale neural network training via compute monitoring, 2023.
\newblock URL \url{https://arxiv.org/abs/2303.11341}.

\bibitem[Ruan et~al.(2023)Ruan, Dong, Wang, Pitis, Zhou, Ba, Dubois, Maddison, and Hashimoto]{Ruan2023}
Yangjun Ruan, Honghua Dong, Andrew Wang, Silviu Pitis, Yongchao Zhou, Jimmy Ba, Yann Dubois, Chris~J. Maddison, and Tatsunori Hashimoto.
\newblock Identifying the risks of lm agents with an lm-emulated sandbox, 2023.
\newblock URL \url{https://arxiv.org/abs/2309.15817}.

\bibitem[Liu et~al.(2023)Liu, Yu, Zhang, Xu, Lei, Lai, Gu, Ding, Men, Yang, Zhang, Deng, Zeng, Du, Zhang, Shen, Zhang, Su, Sun, Huang, Dong, and Tang]{Liu2023}
Xiao Liu, Hao Yu, Hanchen Zhang, Yifan Xu, Xuanyu Lei, Hanyu Lai, Yu~Gu, Hangliang Ding, Kaiwen Men, Kejuan Yang, Shudan Zhang, Xiang Deng, Aohan Zeng, Zhengxiao Du, Chenhui Zhang, Sheng Shen, Tianjun Zhang, Yu~Su, Huan Sun, Minlie Huang, Yuxiao Dong, and Jie Tang.
\newblock Agentbench: Evaluating llms as agents, 2023.
\newblock URL \url{https://arxiv.org/abs/2308.03688}.

\bibitem[Yang et~al.(2023)Yang, Prabhakar, Narasimhan, and Yao]{Yang2023}
John Yang, Akshara Prabhakar, Karthik Narasimhan, and Shunyu Yao.
\newblock Intercode: Standardizing and benchmarking interactive coding with execution feedback, 2023.
\newblock URL \url{https://arxiv.org/abs/2306.14898}.

\bibitem[Askell et~al.(2021)Askell, Bai, Chen, Drain, Ganguli, Henighan, Jones, Joseph, Mann, DasSarma, Elhage, Hatfield-Dodds, Hernandez, Kernion, Ndousse, Olsson, Amodei, Brown, Clark, McCandlish, Olah, and Kaplan]{Askell2021}
Amanda Askell, Yuntao Bai, Anna Chen, Dawn Drain, Deep Ganguli, Tom Henighan, Andy Jones, Nicholas Joseph, Ben Mann, Nova DasSarma, Nelson Elhage, Zac Hatfield-Dodds, Danny Hernandez, Jackson Kernion, Kamal Ndousse, Catherine Olsson, Dario Amodei, Tom Brown, Jack Clark, Sam McCandlish, Chris Olah, and Jared Kaplan.
\newblock A general language assistant as a laboratory for alignment, 2021.
\newblock URL \url{https://arxiv.org/abs/2112.00861}.

\end{thebibliography}

\appendix

\section{A reproducible methodology for calculating the compute-equivalent gain}\label{apA}

This methodology seeks to produce either a lower bound or a point estimate of the CEG of a post-training enhancement In the following, $C(M)$ denotes the training compute of model $M$, and $P(M)$ denotes the performance of model $M$ in a certain task or benchmark.
\begin{enumerate}
\item Pick a paper which tests an enhancement on one or more benchmarks.

\item Identify all benchmarks used to evaluate models \textit{with} and \textit{without} the enhancement.
 For each such benchmark:
\begin{enumerate}
\item Identify all cases where a smaller enhanced model outperforms a larger model without the enhancement. More precisely, identify all pairs of models ML (larger model) and ME (enhanced model) that satisfy these four requirements:
\begin{enumerate}
\item ML is not enhanced
\item ME is enhanced
\item ML was trained using more compute than ME
\item (if possible) ME performs better than ML according to the metric -- $\mathrm{P(ME) >P(ML)}$
\item (if possible) The two models have the same architecture and training method, except for the model size and data size (and the enhancement).
\end{enumerate}

\item The difference in training compute between ME and ML will be used to calculate the CEG. There are several possibilities here:
\begin{enumerate}
\item If all the above requirements are met, the CEG is lower bounded by C(ML) / C(ME).

\item If requirement (iv) can't be satisfied, but P(ML) -- P(ME) is small relative to the overall variation in performance from scaling, then we can use C(ML)/C(ME) as a rough point estimate for the CEG.

\item If requirement (iv) can't be satisfied, but there is enough data at several scales to fit a scaling law to models without the enhancement, the CEG can be estimated directly using this scaling law. See \cref{apA1}.

\item If requirement (v) can't be satisfied, we can try to find a pair such that the architectures and training methods are as similar as possible.\footnote{In some cases, one could use scaling laws to compensate for the difference in architecture, although we did not actually do this.}
\end{enumerate}

\item Keep the highest CEG of all the pairings in the benchmark, unless there are specific reasons to reject some pairs. We came across four such reasons:
\begin{enumerate}
\item If there were multiple non-enhanced baselines, we chose the best non-enhanced baseline, which leads to a lower CEG. This was the case for \hyperref[verification_proc]{verification with process-based feedback}, where we used the best outcome-based baseline.

\item If multiple enhancements are used in a model but we are evaluating the gains from one particular enhancement are of interest, then we discarded the pairs with multiple enhancements. We did this for
  \hyperref[instructgpt]{InstructGPT} and \hyperref[webgpt]{WebGPT}.

\item If the same benchmark is evaluated using fine-tuning datasets with several different sizes, we use the largest dataset for calculating the CEG because we are more interested in more capable models. This was the case for
  \hyperref[verification_proc]{verification with process-based feedback}, where we took the largest dataset size.

\item If the same benchmark is evaluated multiple times with different settings that don't change the CEG, we arbitrarily choose one of them. For example, AlphaCode evaluates a benchmark using different numbers of total samples. We just report the CEG for 10@1K -- and not for 10@10K, 10@100K, and 10@1M -- but a different choice would not have changed the result.
\end{enumerate}

\item Note explicitly any confounders that might lead to over- or under-estimation of the CEG.
\end{enumerate}

\item Note that we don't always report the CEG for all tasks that are presented in a paper, but only those that we believe are most representative of the actual gains. In particular, some confounders and concerns that lead us to reject a task/metric/benchmark are:
\begin{itemize}
\item Using LLM-based evaluation.
\item Benchmarks that measures fairness, alignment, safety or other metrics that are less related to capabilities.
\end{itemize}
\end{enumerate}

\subsection{Compute CEG using a scaling law}
\label{apA1}

\begin{figure}[htb]\centering
 \includegraphics[width=0.45\textwidth]{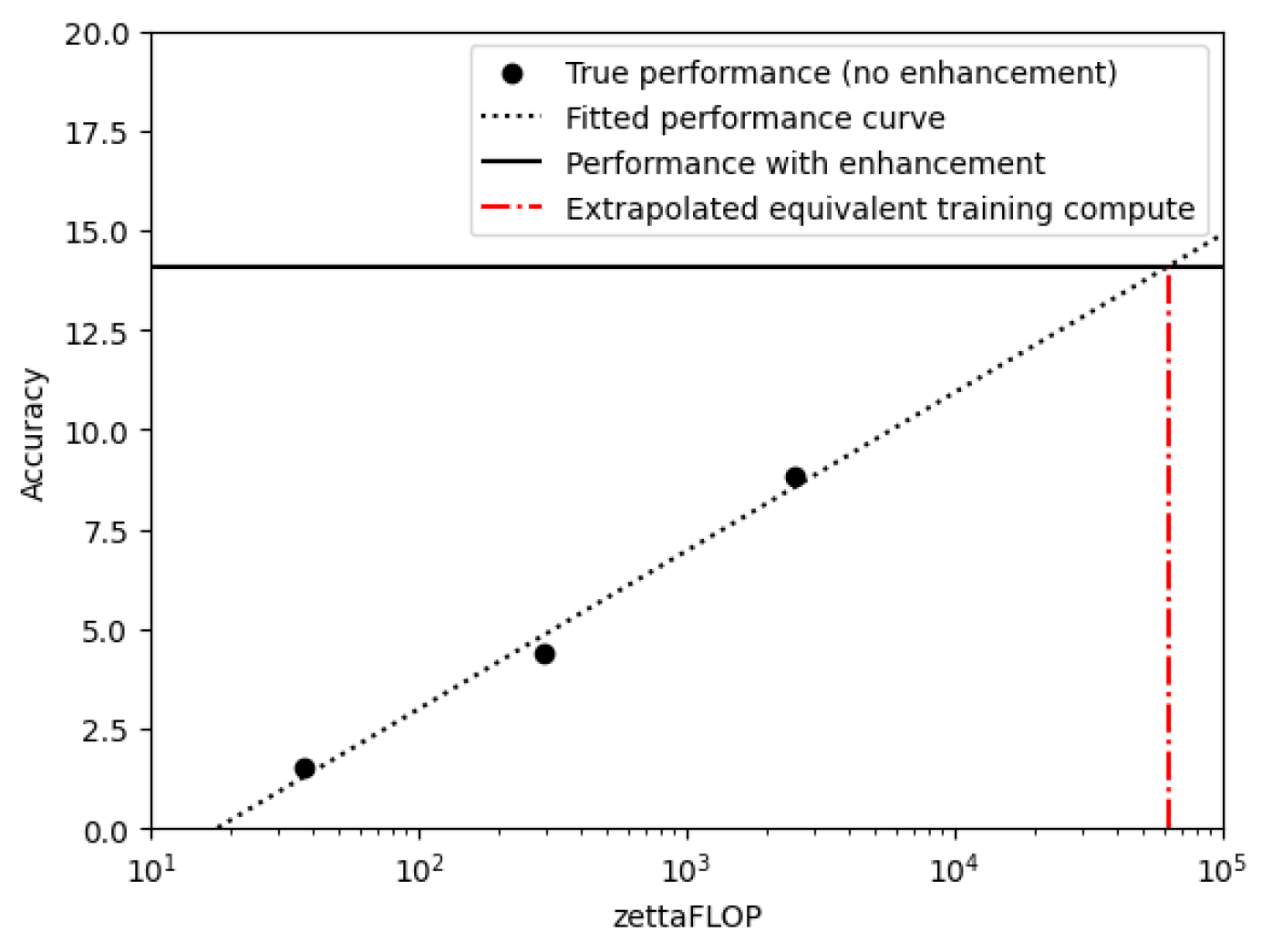}
 \caption{Log-linear scaling model for Minerva.}
 \label{fig:loglinear}
\end{figure}

When the baseline (non-enhanced) architecture is evaluated at at least three different scales, and a simple log-linear or power-law scaling curve fits the data well, we can use this fitted scaling law to estimate the CEG. If the fitted scaling law is $P=f(C)$, then we can approximate the CEG as $f^{-1}(P(ME)) / C(ME)$. An example of this approach for Minerva can be seen in \Cref{fig:loglinear}.\footnote{The code we used is in \href{https://colab.research.google.com/drive/1WgmY7Yk6VaCkzoNmyOkmbpAqwKc7th5a}{this Colab notebook.}}
Note that benchmark performance often exhibits complicated scaling behavior (e.g.: plateaus or inverse scaling), so CEG estimates based on simple scaling laws should be taken with a grain of salt. To make this method more reliable, it would be necessary to precisely characterize the scaling laws for the benchmark, including any nonlinearities, which we have not done.

Fitting performance scaling laws could make it easier to evaluate how performance improvements change with scale, since we could predict how larger models would benefit from an enhancement. We have not done this analysis.

\section{A second estimate of the CEG of chain-of-thought prompting}
\label{apB}

In section 3, we estimated the CEG of Chain of Thought prompting based on data from \citet{Wei2022}. However, this paper reports results from the PaLM family of models, which uses a suboptimal scaling strategy: some of the PaLM models are undertrained. This makes the resulting CEG estimates less reliable.

Therefore, we also estimated the CEG using from Figure 10 in
\citet{Lanham2023}, which presents the accuracy on 8 datasets of language models of different sizes, with and without chain of thought prompting. Although the authors do not provide detailed information about pretraining, we expect these models to be less undertrained than the PaLM models, since the paper dates from long after the improvement in optimal scaling laws from \citet{Hoffmann2022}.

\begin{figure}[htb]\centering
 \includegraphics[width=0.45\textwidth]{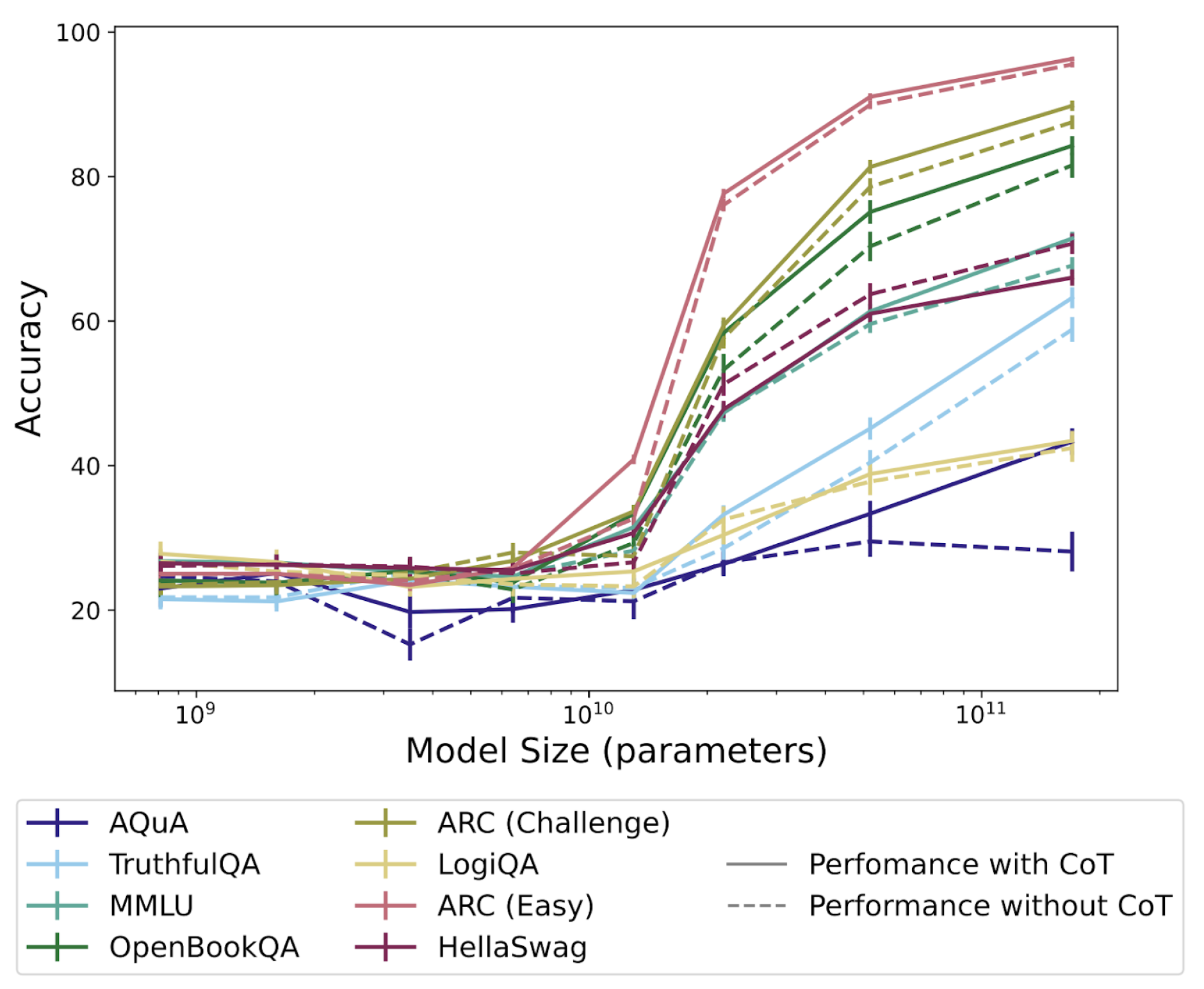}
 \caption*{\textit{Figure 10 from} \citet{Lanham2023}.}
\end{figure}

We estimate the CEG at 52B parameters using the method in Appendix A1:
\begin{itemize}
\item We first fit a log-affine scaling law to the model size
$f(N) = m\log(N) + p$ between 52B and 175B parameters.

\item We then measure the accuracy of ME, the 52B parameter model with chain-of-thought prompting: $P(ME)=Acc$ (52B, with CoT)

\item Finally, we obtain the CEG as the quotient $f^{-1}(P(ME)) / N(ME)$\footnote{This assumes that the 52B and 175B parameter models were trained on the same number of tokens, so that the ratio of model sizes is the ratio of training computes. The paper does not mention this explicitly, but they refer to past Anthropic papers for more information, and models of different size are trained on the same number of tokens in \citet{Askell2021}.}
\end{itemize}

The results can be seen in \Cref{tab:second_cot_estimate}. Overall, these figures are significantly lower than the CEG estimates derived from \citet{Wei2022}. This can be attributed to a few factors. First, the tasks are less symbolic or mathematical, with the exception of Aqua where the chain-of-thought prompting proves to be most beneficial. Second, the models are not as undertrained as those found within the PaLM family.

\begin{table}[h]\centering

\caption{Estimates of the CEG of chain-of-thought prompting using data from \citet{Lanham2023}.}
\label{tab:second_cot_estimate}
\begin{tblr}{colspec={X[1,l]X[1,r]},
column{2}={bg=green!15},
width=0.45\textwidth,
colsep=4.5pt,
rowsep=3pt,
} \hline
Aqua & \SetCell[c=1]{r,m,bg=red!15}0.05 \\
TruthfulQA & 1.4 \\
MMLU & 1.2 \\
OpenBookQA & 1.7 \\
ARC (challenge) & 1.5 \\
LogiQA & 1.3 \\
ARC (Easy) & 1.4 \\
HellaSwag & 2 \\
\hline
\end{tblr}
\end{table}

On the Aqua dataset, the 175B parameter model does \textit{worse} than the 52B parameter model, so our estimate is meaningless. However, chain-of-thought prompting greatly increases the performance of the 175B model (by more than 10 percentage points). This suggests that the CEG at larger model sizes would be even larger on that dataset. However, the benefits of chain-of-thought prompting don't noticeably increase between 52B and 175B on the other datasets.

\end{document}